\documentclass[conference]{IEEEtran}
\usepackage{times}

\usepackage[numbers]{natbib}
\usepackage{multicol}
\usepackage[bookmarks=true]{hyperref}
\usepackage{multirow}

\usepackage{hyperref}
\hypersetup{
	colorlinks=true,
	linkcolor=orange,
	filecolor=magenta,      
	urlcolor=orange,
	citecolor=orange,
}
\usepackage{url}
\usepackage{float}

\usepackage{mathtools}
\usepackage{makecell}
\usepackage{amsfonts}  %
\usepackage{graphics}
\usepackage[pdftex]{graphicx}
\usepackage{wrapfig}
\usepackage{caption}
\usepackage{color}
\usepackage{dsfont}
\usepackage[]{mdframed}
\usepackage{algorithm}
\usepackage[noend]{algpseudocode}
\usepackage{xparse}
\usepackage{amsmath}
\usepackage{bm}
\usepackage{subcaption}
\usepackage{mathtools}
\usepackage{amssymb}
\usepackage{amsthm}
\usepackage{amssymb}
\usepackage{cleveref}

\usepackage{booktabs}
\usepackage{epigraph}

\usepackage{xcolor} %
\usepackage{hyperref} %
\usepackage{booktabs}
\usepackage{multirow}
\usepackage{graphicx}
\usepackage{listings}

\hypersetup{
    colorlinks=true,
}

\usepackage{colortbl} %
\definecolor{lightblue}{RGB}{217, 237, 247} %
\definecolor{lightgreen}{RGB}{229, 245, 224}

\usepackage{xargs} %
\usepackage[colorinlistoftodos,prependcaption,textsize=tiny]{todonotes}
\newcommandx{\wrn}[2][1=]{\todo[linecolor=red,backgroundcolor=red!25,bordercolor=red,#1]{#2}}
\newcommandx{\cmt}[2][1=]{\todo[linecolor=blue,backgroundcolor=blue!25,bordercolor=blue,#1]{#2}}

\algrenewcommand{\algorithmiccomment}[1]{\hfill\textcolor{lightgray}{$\triangleright$\;#1}}

\renewcommand{\cref}{\Cref}  %

\usepackage{listings}
\usepackage{xcolor}
\lstdefinelanguage{Python}{
  morekeywords=[1]{True, False, None, and, as, assert, async, await, break, class, continue, def, del, elif, else, except, finally, for, from, global, if, import, in, is, lambda, nonlocal, not, or, pass, raise, return, try, while, with, yield},
  morekeywords=[2]{transformers, numpy},
  sensitive=true,
  morecomment=[l]{\#},
  morestring=[b]',
  morestring=[b]",
}

\definecolor{darkgreen}{RGB}{0,128,0}
\definecolor{darkblue}{RGB}{0,0,255}
\definecolor{darkred}{RGB}{139,0,0}

\def \ModelAcronym {FAST}
\def \ModelUniversalAcronym {FAST+}
\def \ModelAcronymLong {\underline{\textbf{F}}requency-space \underline{\textbf{A}}ction \underline{\textbf{S}}equence \underline{\textbf{T}}okenization (\textbf{FAST})}
\def \GeneralistModelAcronym {$\pi_0$-FAST}

\IEEEoverridecommandlockouts

\begin{document}

\title{
\ModelAcronym: Efficient Action Tokenization for Vision-Language-Action Models
}

\pdfinfo{
   /Author (Physical Intelligence)
   /Title  (@title)
   /Subject (Robot Foundation Models)
   /Keywords (Robot Foundation Models)
}

\def\cameraready{0}  %

\ifx\cameraready\undefined
    \author{
    Anonymous Submission
    }
\else
    \author{
    Karl Pertsch$^{\ast, 1, 2, 3}$, Kyle Stachowicz$^{\ast, 2}$,\\[0.1cm]
    Brian Ichter$^{1}$, Danny Driess$^{1}$, Suraj Nair$^{1}$, Quan Vuong$^{1}$, Oier Mees$^{2}$, Chelsea Finn$^{1, 3}$, Sergey Levine$^{1, 2}$\\[0.3cm]
    
    $^{1}$Physical Intelligence, $^{2}$UC Berkeley, $^{3}$Stanford\\[0.2cm]
    \thanks{$^\ast$: Core contributors\newline Correspondence to: \texttt{research@physicalintelligence.company}}
    \url{https://pi.website/research/fast}
    }
\fi

\maketitle

\begin{abstract}

Autoregressive sequence models, such as Transformer-based vision-language action (VLA) policies, can be tremendously effective for capturing complex and generalizable robotic behaviors. 
However, such models require us to choose a tokenization of our continuous action signals, which determines how the discrete symbols predicted by the model map to continuous robot actions.
We find that current approaches for robot action tokenization, based on simple per-dimension, per-timestep binning schemes, typically perform poorly when learning dexterous skills from high-frequency robot data. 
To address this challenge, we propose a new compression-based tokenization scheme for robot actions, based on the discrete cosine transform. Our tokenization approach, \ModelAcronymLong, enables us to train autoregressive VLAs for highly dexterous and high-frequency tasks where standard discretization methods fail completely. Based on \ModelAcronym{}, we release \ModelUniversalAcronym, a \emph{universal} robot action tokenizer, trained on 1M real robot action trajectories. It can be used as a black-box tokenizer for a wide range of robot action sequences, with diverse action spaces and control frequencies. Finally, we show that, when combined with the $\bm{\pi_0}$ VLA, our method can scale to training on 10k hours of robot data and match the performance of diffusion VLAs, while reducing training time by up to 5x.

\end{abstract}

\IEEEpeerreviewmaketitle

\section{Introduction}
\label{sec:intro}

\begin{figure}[]
    \centering
    \vspace{-0.5cm}
    \includegraphics[width=\linewidth]{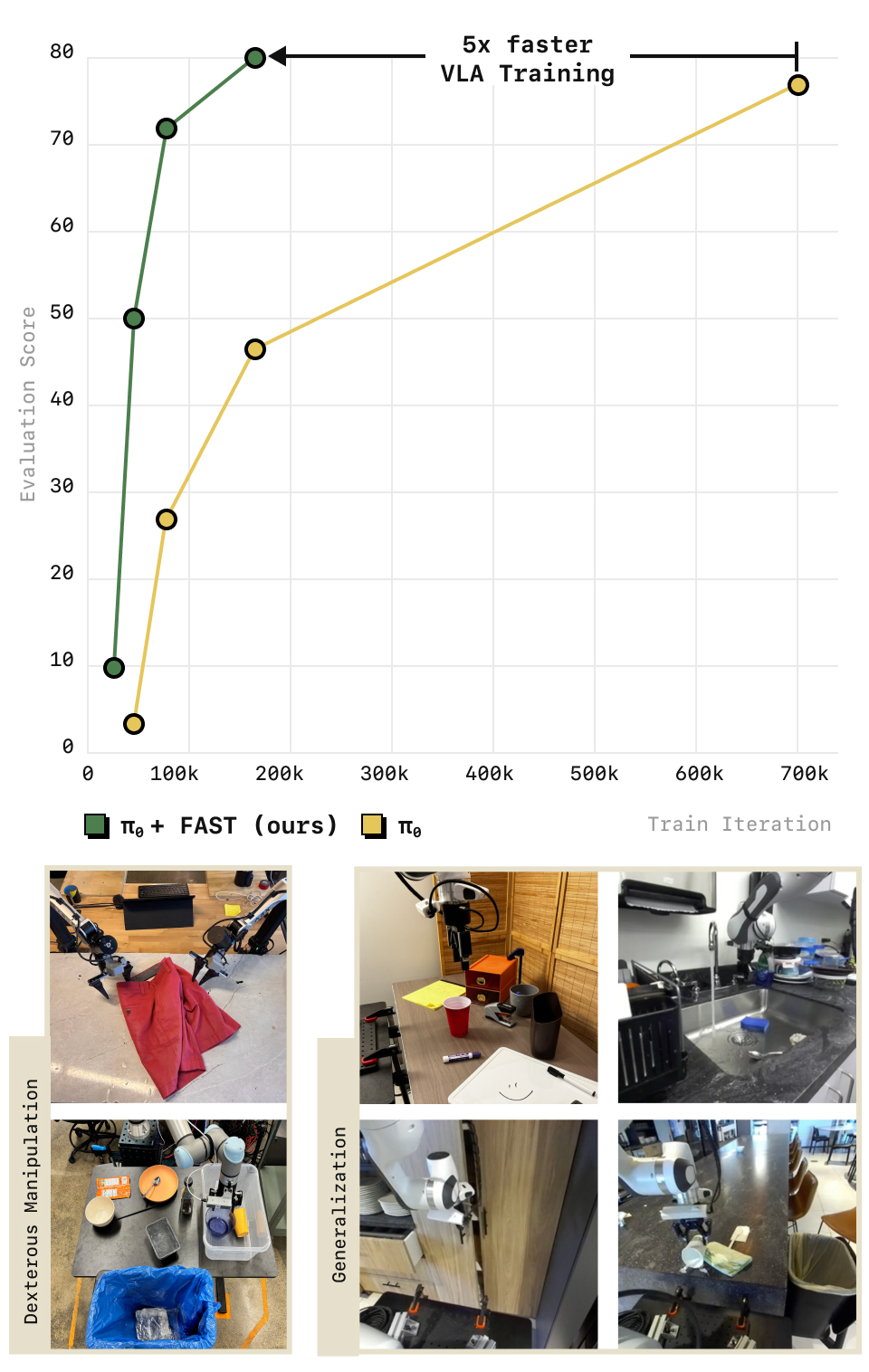}
    \caption{We propose \ModelAcronym, a simple yet effective approach for tokenization of robot action trajectories via time-series compression. \ModelAcronym{} enables training of autoregressive VLAs that solve complex dexterous manipulation tasks and generalize broadly to new scenes. We use it to train \GeneralistModelAcronym{}, a generalist robot policy that matches the performance of the state-of-the-art $\pi_0$ diffusion VLA on dexterous and long-horizon manipulation tasks, while training 5x faster (\textbf{top}).
    }
    \label{fig:convergence}
    \vspace{-1.5em}
\end{figure}

\begin{figure}[t]
    \centering
    \includegraphics[width=\linewidth]{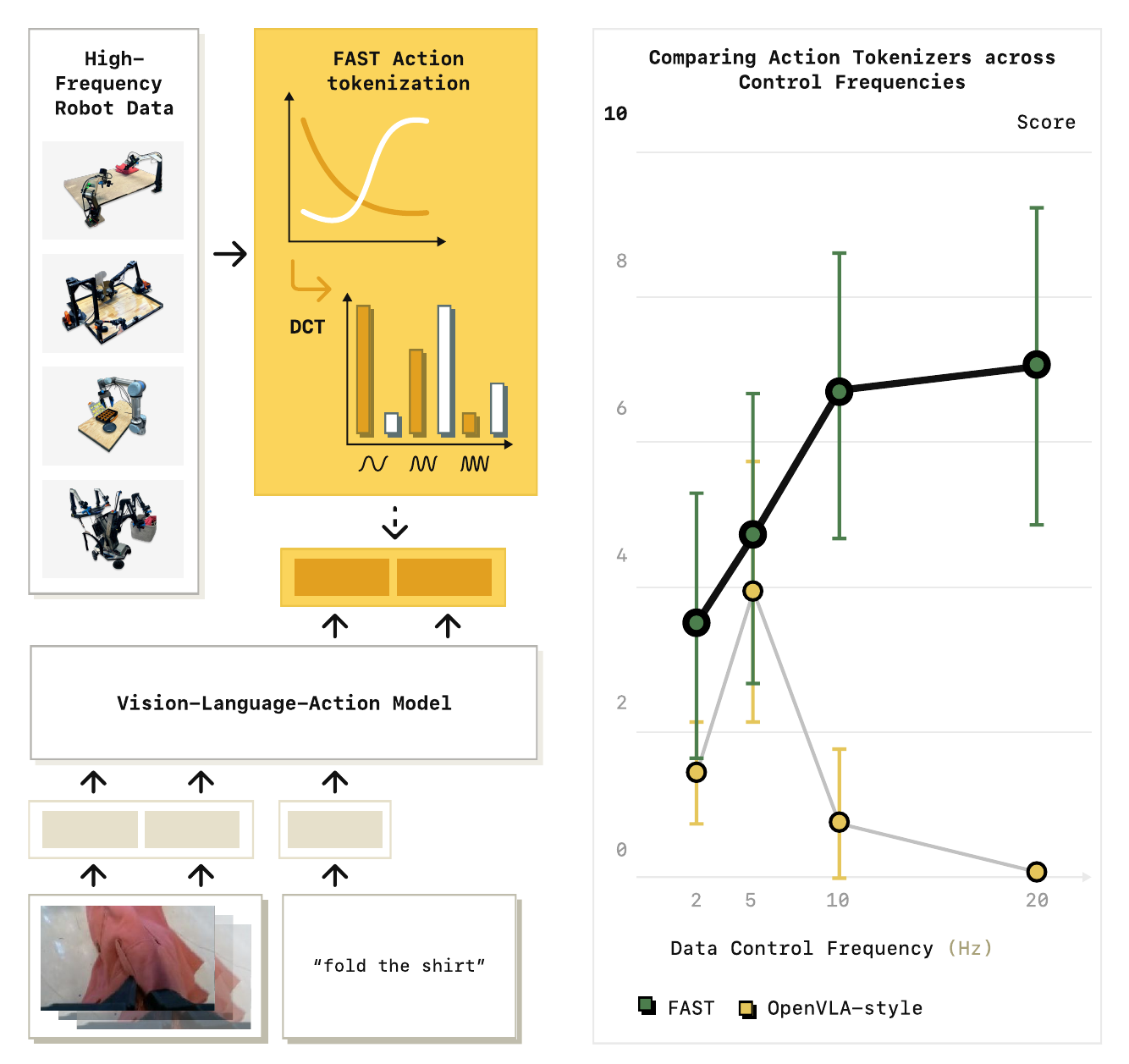}
    \caption{\textbf{Left}: \ModelAcronym{} tokenization enables training of autoregressive Transformers for dexterous robot control via simple next token prediction. \textbf{Right}: \ModelAcronym{} outperforms popular binning tokenization schemes, e.g., used in OpenVLA~\citep{kim2024openvla}, particularly for high-frequency robot data.
    }
    \label{fig:teaser}
\end{figure}

Large, high-capacity Transformer models can be tremendously effective for capturing complex and generalizable robotic behaviors both from scratch~\citep{rt12022arxiv,zhao2023learning,octo_2023,bharadhwaj2023roboagent,Doshi24-crossformer, wangscaling} and using models pre-trained for next-token prediction on Internet-scale image-text corpora~\citep{rt22023arxiv,kim2024openvla,wen2024tinyvlafastdataefficientvisionlanguageaction,black2024pi_0,ye2024latent}.
However, these models require choosing a tokenization of the continuous action signal, which determines how the discrete symbols predicted by the model map to continuous robot actions~\citep{yan2024elastictok,jang2024efficient,lee2024behavior,chen2022beats}. It is widely known that a good choice of tokenization can be critical to the performance of sequence models~\citep{radford2019language, sennrich2015neural}. Prior robotic policies of this sort typically use na\"{i}ve tokenization strategies based on a per-dimension, per-timestep binning scheme~\citep{brohan2022rt,rt22023arxiv,kim2024openvla}.
We find that such methods perform poorly when learning dexterous skills with high-frequency control (see \cref{fig:teaser}, right). We observe that correlations between time steps are a major challenge for na\"{i}ve tokenization strategies when predicting sequences of future actions, i.e., action ``chunks'', as is common for high-frequency control. Highly correlated action tokens \emph{diminish} the effectiveness of the next token prediction objective used in autoregressive VLAs. Intuitively, in such cases low token prediction loss can often be achieved with mappings as trivial as simply copying the most recent action token, leaving models in poor local optima.

In this work, we propose a new tokenization strategy from first principles. Our key insight is that robot action signals need to be \emph{compressed} before training, to reduce correlation between consecutive tokens. We take inspiration from compression-based tokenization strategies, such as the byte-pair encoding method commonly used by language models~\citep{gage1994new, sennrich2015neural}. However, since robotic actions are continuous, the corresponding compression strategy should be chosen accordingly. We therefore base our method off of the discrete cosine transform~(DCT) encoding, which is widely used for compressing continuous signals such as images (e.g., JPEG compression).
We find that the resulting tokenization approach, \ModelAcronymLong, enables us to train autoregressive VLA policies via simple next token prediction (see \cref{fig:teaser}, left) for highly dexterous and high-frequency tasks where standard discretization methods fail entirely. 
Additionally, \ModelAcronym{} for the first time enables efficient VLA training on the recently introduced DROID dataset~\citep{khazatsky2024droid}, a large-scale multitask ``in-the-wild'' robot manipulation dataset. The resulting policy is the first language-conditioned generalist manipulation policy that can be successfully evaluated \emph{zero-shot} in unseen environments, simply by prompting it in natural language.

Based on \ModelAcronym{}, we develop \ModelUniversalAcronym, a \textbf{\underline{universal} robot action tokenizer}, trained on 1M real robot action trajectories that cover a large diversity of robot embodiments, action spaces and control frequencies. We demonstrate that the \ModelUniversalAcronym~tokenizer effectively tokenizes a wide range of robot action sequences, from single-arm to bi-manual and mobile robots, and is a good off-the-shelf tokenizer for training autoregressive VLA models. When integrated with the $\pi_0$ VLA, FAST-based autoregressive VLAs scale to training on 10k hours of robot data and achieve performance comparable to diffusion-based VLAs across a variety of tasks, while reducing training time by up to 5x (see \cref{fig:convergence}).

\section{Related Work}
\label{sec:related}

\noindent\textbf{Tokenization for language, text, and audio.}
Tokenization is a key component of training pipelines for modern transformer-based autoregressive sequence models, and the choice of tokenization approach can have significant impact on model training and downstream performance~\cite{radford2019language}. While there are multiple works exploring the training of ``tokenization-free'' language models \citep{gillick2016bytelm, meta_blt} that directly operate on bit streams, most language models today rely on a text tokenization stage prior to training. 
A common approach is byte pair encoding~\citep{gage1994new, radford2019language}, which compresses input text by merging frequently occurring token sequences into new tokens. For images, \emph{learned} compression schemes present an effective approach: input images can be represented as ``soft tokens'' produced by a pre-trained vision encoder~\citep{liu2023llava}, and full autoregressive image input-output can be achieved with a vector-quantizing autoencoder~\cite{esser2020taming, vqvae}. Similar approaches can be extended to the video domain~\cite{yu2023magvit}. In audio generation and speech synthesis, which share the time-series structure of action prediction, state-of-the-art models typically encode time-series audio data using either frequency-domain spectrogram images~\citep{gong21b_interspeech} or using learned vector quantizers~\citep{zeghidour2021soundstreamendtoendneuralaudio}.

\noindent\textbf{Vision-language-action models.}
Recently, multiple works have developed \emph{generalist} robot policies~\citep{brohan2022rt,octo_2023,bharadhwaj2023roboagent,rt22023arxiv,Doshi24-crossformer,kim2024openvla,wangscaling,cheang2024gr2generativevideolanguageactionmodel}
that are trained on increasingly large robot learning datasets~\citep{open_x_embodiment_rt_x_2023,khazatsky2024droid,walke2023bridgedata,fang2024rh20t,mandlekar2018roboturk,jiang2024dexmimicgen}. 
One promising approach for training generalist policies are vision-language-action models (VLAs; ~\citep{rt22023arxiv,collaboration2023open,kim2024openvla,Zawalski24-ecot,black2024pi_0,wen2024tinyvlafastdataefficientvisionlanguageaction,zheng2024tracevla,zhen20243d,cheng2024navila,cheang2024gr2generativevideolanguageactionmodel}).
VLAs fine-tune vision-language models, that are pre-trained on internet-scale image and text data, for robot control. This has multiple benefits: using large vision-language model backbones, with billions of parameters, provides policies with the necessary expressivity for fitting large robot datasets. Reusing weights pre-trained on internet-scale datasets also improves the ability of VLAs to follow diverse language commands and generalize, e.g., to new objects and scene backgrounds~\citep{rt22023arxiv,kim2024openvla,Zawalski24-ecot,wen2024tinyvlafastdataefficientvisionlanguageaction,jones2025beyond}. 
Most VLA models today are confined to rather simple, low-frequency control tasks, particularly models that use the most common autoregressive VLA design~\citep{rt22023arxiv, kim2024openvla}. We show that this is a direct consequence of the \emph{action tokenization} schemes employed by these models, which make training on dexterous tasks challenging. We introduce a new action tokenization approach that allows us to train the first autoregressive VLAs on dexterous and high-frequency robot data.

\noindent\textbf{Action representations for VLA training.}
Prior works have explored various action parameterizations for training robot policies, including VLAs.
One line of work uses ``semantic'' action representations like language sub-tasks~\citep{driess2023palm,saycan2022arxiv,belkhale2024rthactionhierarchiesusing}, 
or keypoints~\citep{nasirianypivot,huang2024rekep,fangandliu2024moka,dipalo2024kat}. 
Such approaches can often learn from few examples or even perform tasks \emph{zero-shot} without any robot examples~\citep{nasirianypivot,huang2024rekep,fangandliu2024moka}, but require hand-designed low-level controllers for task execution, limiting their generality. An alternative approach directly trains VLAs to output low-level robot control commands given image and language instruction inputs. The most common design directly embeds actions into discrete tokens, that can be generated with standard autoregressive sequence models, like any popular vision-language model. Existing approaches map from continuous robot actions to discrete action tokens using a simple per-dimension, per-timestep binning scheme~\citep{brohan2022rt,rt22023arxiv,kim2024openvla}.
We find that this scheme struggles to scale to high-frequency robot control tasks. We propose a new tokenization scheme for robot actions, based on time-series compression techniques, that allows us to train autoregressive VLAs on high-frequency data. A number of works have also proposed alternatives to tokenization, for example by using regression heads or introducing new weights for diffusion decoding~\citep{Doshi24-crossformer,black2024pi_0,lee2024behavior,wen2024tinyvlafastdataefficientvisionlanguageaction}. In comparison, our approach does not require modifications of the underlying pre-trained transformer model, can easily be applied to any pre-trained autoregressive transformer model, and achieves competitive performance to state-of-the-art diffusion-based VLAs~\citep{black2024pi_0} across many tasks, while being significantly more compute efficient to train. 

Another set of related work explores \textit{vector-quantized} action representations~\citep{lee2024behavior, belkhale2024minivla, mete2024questselfsupervisedskillabstractions}. Such approaches train a vector-quantized encoder-decoder network, for which reconstruction quality can be sensitive to hyperparameter choices and structure~\cite{yu2023magvit}. We find that these methods perform well at coarse, low-fidelity reconstruction tasks, but fail on high-frequency tasks when fine-grained control is required. In comparison, our {\ModelAcronym} tokenization scheme has few hyperparameters and can reconstruct actions with high precision while offering strong compression properties.

\section{Preliminaries}
\label{sec:prelim}

\noindent\textbf{Problem formulation.}
Our goal is to train policies $\pi(a_{1:H} \vert o)$ that map an observation $o$ to a sequence of future robot actions $a_{1:H}$. We assume that policies output an ``action chunk''~\cite{zhao2023learning,laiaction}, a \emph{sequence} of $H$ actions~\citep{chi2023diffusionpolicy,black2024pi_0,zhao2023learning}, which makes it easier to produce temporally-consistent actions and reduces compounding error. The goal of \textit{action tokenization} is to define a mapping $\mathcal{T}_a: a_{1:H} \rightarrow [T_1, \dots, T_n]$
from a sequence of continuous actions $a_{1:H}$, with dimensionality $\vert \mathcal{A}\vert$, to a sequence of $n$ discrete tokens $T \in \vert \mathcal{V}\vert$ from a vocabulary of size $\vert \mathcal{V}\vert$. Note that the number of tokens $n$ may differ between action sequences, just like sentences of the same length may be tokenized into a variable number of text tokens. 

\noindent\textbf{Binning-based action tokenization.}
The most commonly used approach for action tokenization is a simple binning discretization scheme~\citep{rt12022arxiv,rt22023arxiv,kim2024openvla,zhen20243dvla,reed2022generalist}. For a given action $a$, this approach discretizes each dimension independently, dividing the range of values in the training dataset into $N$ uniform bins, most commonly using $N=256$. For a \emph{sequence} of $D$-dimensional actions $a_{1:H}$, this tokenization scheme would be applied to each time step, resulting in a final token sequence $\mathcal{T}_a\big(a_{1:H}\big) = [T_{1, 1}, \dots, T_{1, D}, \dots, T_{H, 1}, \dots, T_{H, D}]$. For high-frequency robot data, this tokenization scheme is sub-optimal: it can easily produce hundreds of tokens per action chunk, which make training challenging and lead to slow inference.

\section{Case Study: How Does Tokenization Affect VLA Training?}
\label{sec:educational_example}

\begin{figure}
    \centering
    \includegraphics[width=\linewidth]{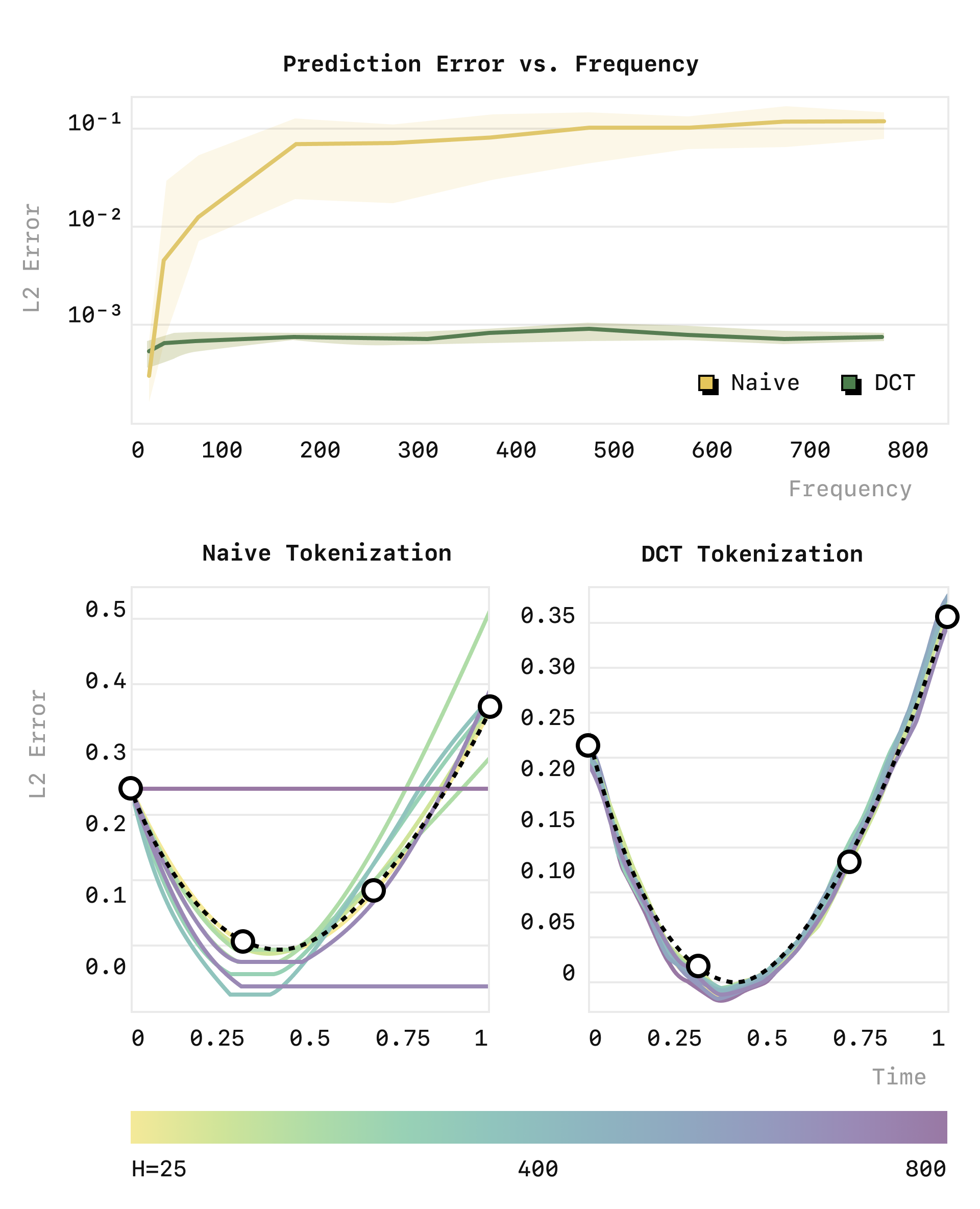}
    \caption{\textbf{Effect of sampling rate on prediction performance.} We train a small autoregressive transformer model on a didactic interpolation task, in which the network must predict the black dashed curve given the four circles. 
    We find that models trained with the binning tokenization approach used in prior VLAs~\cite{rt22023arxiv, kim2024openvla} produce increasingly poor predictions as we increase the sampling frequency of the underlying signal, due to strong correlation between consecutive tokens at high frequencies. Our \ModelAcronym{} tokenization approach, based on the discrete cosine transform (DCT), addresses the problem and leads to high-quality predictions across all sampling rates.
    }
    \label{fig:toy_example}
\end{figure}

To illustrate the challenge of training autoregressive policies with current action tokenization approaches, we start with a simple didactic example. We create a synthetic time-series dataset where the goal is to predict a cubic spline that interpolates four randomly-generated points (see \cref{fig:toy_example}, bottom). This toy problem reflects the challenge faced by policies trained on high-frequency action chunks, which must predict a sequence of continuous actions given some conditioning information. We tokenize the target sequences using the na\"{i}ve tokenization scheme employed in previous VLA policies, which discretizes each element in the sequence separately into one of 256 bins (see \cref{sec:prelim}). We then train a small, autoregressive transformer policy to predict the tokenized signal given the conditioning points. We repeat this experiment for different \emph{sampling rates} of the target signal, from 25 to 800 timesteps per sequence, without changing the underlying dataset. This emulates training autoregressive policies on action data collected at different frequencies.

The average prediction MSE of autoregressive models trained at different frequencies is shown in \cref{fig:toy_example}, top (``naive'').
We observe that the model with binning tokenization achieves good prediction performance (i.e., low MSE) for low sampling rates. But as the sampling rate increases, the prediction error steeply increases, until eventually the model simply copies the first action, as seen in the qualitative visualization in \cref{fig:toy_example}, bottom left. Note that this issue \emph{cannot} be attributed to the data itself: the complexity of the underlying data distribution does not change, and we would expect a model with the same capacity trained for the same number of steps to achieve comparable performance across all sampling rates. %
So what happened?

To understand how the tokenization scheme impacts learning performance, we need to look at the learning objective itself. Fundamentally, autoregressive models are trained to predict the next token, given all previous tokens. As such, their learning signal is proportional to the marginal information content of $T_i$ given $T_{1:i-1}$. Crucially, when using the na\"{i}ve per-timestep tokenization scheme, this marginal information \emph{approaches zero} as the control frequency of the training signal increases: for smooth signals, as timesteps get shorter the change per timestep decreases proportionally.
This greatly \emph{slows down} the rate of convergence during training and can make it challenging to fit complex, high-frequency datasets. Indeed, such challenges have been observed in prior work. For instance, OpenVLA worked well on the low-frequency BridgeV2 and RT-1 datasets, but has struggled to fit the higher-frequency DROID dataset~\citep{kim2024openvla}. The result of our case study underlines the importance of designing better tokenization schemes for robot actions.

\section{Efficient Action Tokenization via Time-Series Compression}
\label{sec:method}

\begin{figure*}[t]
    \centering
    \includegraphics[width=\linewidth]{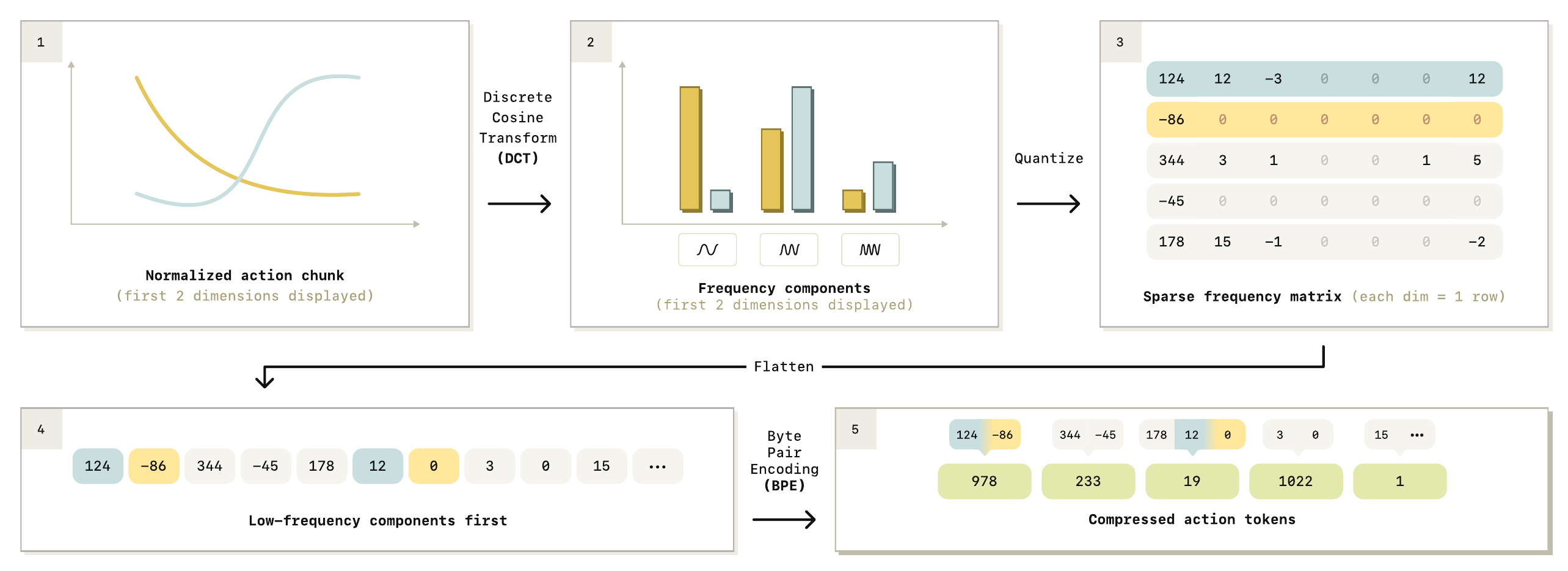}
    \caption{\textbf{Overview of the \ModelAcronym~action tokenization pipeline.} Given a normalized chunk of actions, we apply discrete cosine transform (DCT) to convert the signal to the frequency domain. We then quantize the DCT coefficients and use byte-pair encoding (BPE) to compress the flattened sequence of per-dimension DCT coefficients into the final action token sequence. See \cref{sec:dct_tokenizer} for a detailed description.}
    \label{fig:method_overview}
\end{figure*}

We saw in the previous section how redundancy in high-frequency action trajectories can lead to low marginal information for each action token, and thereby poor training performance. To address this, we need a tokenization approach that compresses the highly redundant action signal into a smaller number of high-information tokens. In this section, we will first describe a simple approach for compressing continuous time series (\ref{sec:dct}), then use it to design an action tokenization algorithm (\cref{sec:dct_tokenizer}), and finally explain how we train a \emph{universal} tokenizer for robot actions (\cref{sec:universal_tokenizer}).

\subsection{Time-Series Compression via Discrete Cosine Transform}
\label{sec:dct}

There is a rich body of work on effectively compressing continuous time series, from approaches that compress signals after transforming them into the frequency domain \citep{fft, dct, jpeg} to \emph{learned} compression approaches, e.g., based on vector quantization~\cite{vqvae,fsq}. One key takeaway of our work is that \emph{any} sufficiently effective compression approach, when applied to the action targets, is suited to improve the training speed of VLA models. In practice, there are a few considerations that may still lead us to favor some compression algorithms over others, e.g., the complexity of training the tokenizer, and how efficient is it at tokenizing and detokenizing actions.

In this work, we use a compression algorithm based on the discrete cosine transform (DCT)~\cite{dct}. DCT is a frequency-space transform that represents a continuous signal as a sum of cosine elements of various frequencies. Low frequencies capture the overall shape of the signal, while high-frequency components reflect sharp jumps. DCT is a commonly used transformation for compression algorithms, e.g., for JPEG image compression~\citep{jpeg}, due to its simplicity and computational efficiency, and its strong compression property on practical images: since pixels often vary smoothly, DCT can often represent most of the information of an input signal in only a few coefficients. Signals can be compressed by omitting frequency components with low weights. Compared to learned compression approaches based on vector quantization, DCT-based compression is an analytical approach, thus extremely simple and fast.

\subsection{The \ModelAcronym~Tokenization Algorithm}
\label{sec:dct_tokenizer}

We use the discrete cosine transform to design \ModelAcronym, a quick and effective tokenization approach for robot actions. We detail the steps from raw robot actions to action tokens in \cref{fig:method_overview}. We first normalize the input actions, such that the 1st and 99th quantile of values in the training dataset for each action dimension maps to the range $[-1, \dots, 1]$. This initial normalization step is useful to bring the data into a specified range and also makes tokenization of cross-embodied datasets with different action scales easier. We use quantiles to be robust to outlier actions which occasionally occur in large robot datasets. After the data is normalized, we apply the discrete cosine transform to each action dimension separately. To compress the DCT-converted signal we can simply omit insignificant coefficients, which we implement through a scale-and-round operation, where the scaling coefficient is a hyperparameter that trades off between lossiness and compression rate of the tokenization operation.

After the rounding operation, the DCT coefficient matrix is typically sparse, with most entries being zero and only a few significant coefficients remaining per action dimension. To actually realize the compression, we must convert this sparse matrix into a sequence of dense tokens. We flatten the matrix into a 1-dimensional vector of integers, interleaving action dimensions by including all low-frequency components first, and train a byte pair encoding (BPE) tokenizer~\citep{gage1994new} to losslessly compress it into dense action tokens. The BPE step ``squashes'' the zero-valued components and merges frequently-occurring coefficient combinations across action dimensions. We choose BPE to compress the DCT matrix, since many efficient implementations exist and it can produce a fixed-size output vocabulary that can be easily integrated into the existing vocabulary of vision-language models for VLA training. Other lossless compression algorithms like Huffman coding~\citep{huffmancode} or Lempel-Ziv methods~\citep{lempelziv} (the algorithms underlying the gzip compression approach) could be used instead, but we leave this investigation for future work.

Note that the \emph{order} of flattening the $\vert A \vert \times H$ DCT coefficient matrix prior to BPE encoding can have significant impact on policy training. There are two options: column-first flattening, i.e., concatenate the lowest-frequency components for each dimension first, or row-first flattening, i.e., concatenating all frequency components for a single action dimension first. We choose the former, since we find that predicting the \emph{low-frequency} components, that characterize the overall shape of the output sequence, first during autoregressive prediction leads to more stable policy rollouts.

\begin{algorithm}[t]

\caption{\ModelAcronym~Tokenizer}
\label{alg:fast}
\begin{algorithmic}
\Require scale $\gamma$, (for inference) BPE dictionary $\Phi$
\Procedure{FASTTokenizer}{$a_{1:H}$}
    \State $C^i_{j} \gets \texttt{DCT}\left(a^{i}_{1:H}\right)$ \Comment{Compute DCT coefficients}
    \State $\bar C^i_{j} \gets \texttt{round}\left(\gamma \cdot C^i_{j}\right)$ \Comment{Quantize coefficients}
    \State $\left[T_k\right] \gets \left[\bar C^1_1, \bar C^2_1, \dots, C^1_2, \dots, C^n_H\right]$ \Comment{Flatten tokens}

\noindent \textbf{BPE Training}:
    \State $\phi \gets \texttt{TrainBPE}(\mathcal{D} := \{[T_k]\})$

\noindent \textbf{Tokenization}:
    \State $\left[{\bar T}_1, \dots, {\bar T}_{\bar k}\right] \gets \texttt{BPE}\left([T_1, \dots, T_k], \phi\right)$
    \State \Return $\text{action\_tokens}$
\EndProcedure
\end{algorithmic}
\end{algorithm}

All operations in our tokenization pipeline are easily invertible, allowing fast decoding of predicted actions. The tokenizer has only two hyperparameters: the scale applied to the DCT coefficients before rounding, and the vocabulary size of the BPE compression step. We find that both parameters are not very sensitive, and we use the same values across all our single-dataset tokenization experiments (rounding scale 10, BPE vocabulary size 1024). This is in contrast to end-to-end \emph{learned} compression modules that rely on vector quantization~\cite{vqvae}. Such networks are often tedious to train, and require careful dataset-specific hyperparameter selection to achieve good reconstruction~\cite{yu2023magvit,fsq}. Our experiments show that our DCT-based tokenization approach trains higher-performing policies than VQ-based approaches, while being significantly simpler and easier to tune. %

We empirically demonstrate the benefits of our DCT-based tokenization in the toy example from \cref{sec:educational_example}. \cref{fig:toy_example} shows that training the autoregressive model on DCT-compressed target tokens achieves constantly low prediction error across a wide range of sampling frequencies. We provide a concise summary of our tokenization approach in \cref{alg:fast} and test the effectiveness of \ModelAcronym~tokenization on robot control problems in \cref{sec:experiments}.

\subsection{A Universal Robot Action Tokenizer}
\label{sec:universal_tokenizer}

The only \emph{learned} component of our tokenizer is the vocabulary of the BPE encoder, which needs to be trained for each new dataset that the tokenizer is being applied to. While this learning process is fast (typically only a few minutes), it adds additional friction to using \ModelAcronym~tokenization. Thus, we aim to train a \textbf{universal} action tokenizer, that can encode chunks of robot actions from \emph{any} robot. To this end, we train a tokenizer using the pipeline described above on a large, cross-embodied robot action dataset, consisting of approximately one million 1-second action chunks from single-arm, bi-manual and mobile manipulation robots, with joint and end-effector control action spaces and various control frequencies. We provide a detailed breakdown of the data mixture used for training the universal tokenizer in \cref{sec:app_universal_data_mix}. Once trained, our universal action tokenizer, \ModelUniversalAcronym, can be applied as a black-box tokenizer on 1-second action sequences from any robot setup. Our experimental evaluation shows that it is competitive to tokenizers tuned for individual datasets. 

\noindent\textbf{Code release.} We release our pre-trained universal action tokenizer, \ModelUniversalAcronym, in a convenient HuggingFace \texttt{AutoProcessor} class, that makes it easy to apply the tokenizer to any new robot action chunk in three lines of code:

\lstset{
  language=Python,
  basicstyle=\ttfamily\small,
  keywordstyle=[1]\color{green!50!black}\bfseries,   %
  keywordstyle=[2]\color{blue}\bfseries,   %
  commentstyle=\color{grey},                      %
  stringstyle=\color{red!70!black},               %
  identifierstyle=\color{black},       %
  showstringspaces=false,
}

\begin{lstlisting}[language=Python]
from transformers import AutoProcessor

tokenizer = AutoProcessor.from_pretrained(
    "physical-intelligence/fast", 
    trust_remote_code=True
)
tokens = tokenizer(action_chunk)
\end{lstlisting}

For best compression results, we recommend normalizing input actions to range $[-1, \dots, 1]$ via quantile normalization as described in \cref{sec:dct_tokenizer}, and tokenizing 1-second action chunks at a time. 
Our module also makes it easy to train a \emph{new} \ModelAcronym~tokenizer on a given dataset of action chunks:
\begin{lstlisting}[language=Python]
from transformers import AutoProcessor

tokenizer = AutoProcessor.from_pretrained(
    "physical-intelligence/fast", 
    trust_remote_code=True
)
new_tokenizer = tokenizer.fit(action_dataset)
\end{lstlisting}

\section{Experiments}
\label{sec:experiments} 

In our experiments, we test \ModelAcronym{} with two VLA backbones: $\pi_0$~\citep{black2024pi_0} and OpenVLA~\citep{kim2024openvla}. We compare \ModelAcronym{} to alternative action tokenization schemes and ablate key design decisions. We then compare $\pi_0$ models trained with \ModelAcronym{} tokenization to the state-of-the-art $\pi_0$ flow-matching (diffusion) VLA, and test the scaling of autoregressive VLA training with \ModelAcronym{} to large, cross-embodied datasets with 10k hours of dexterous robot manipulation data.

\subsection{Experimental Setup}
\label{sec:exp_setup}

\noindent\textbf{Policy implementation.}
We test different tokenization schemes for autoregressive VLA training with popular VLA backbones. For most of our experiments, we use $\pi_0$~\citep{black2024pi_0}, a VLA based on PaliGemma-3B~\citep{beyer2024paligemma}. We also test with OpenVLA~\citep{kim2024openvla}, which is built on Prismatic~7B~\citep{karamcheti2024prismatic}. During training, we tokenize 1-second action chunks and overwrite the least used tokens in the VLM vocabulary with the resulting action tokens, following prior VLAs~\citep{rt22023arxiv,kim2024openvla}. We fine-tune the VLA models for robot action prediction, without weight freezing. %
We provide more details on the policy training setup in \cref{sec:app_policy_training_details}.

\begin{figure}[t]
    \centering
    \includegraphics[width=0.9\linewidth]{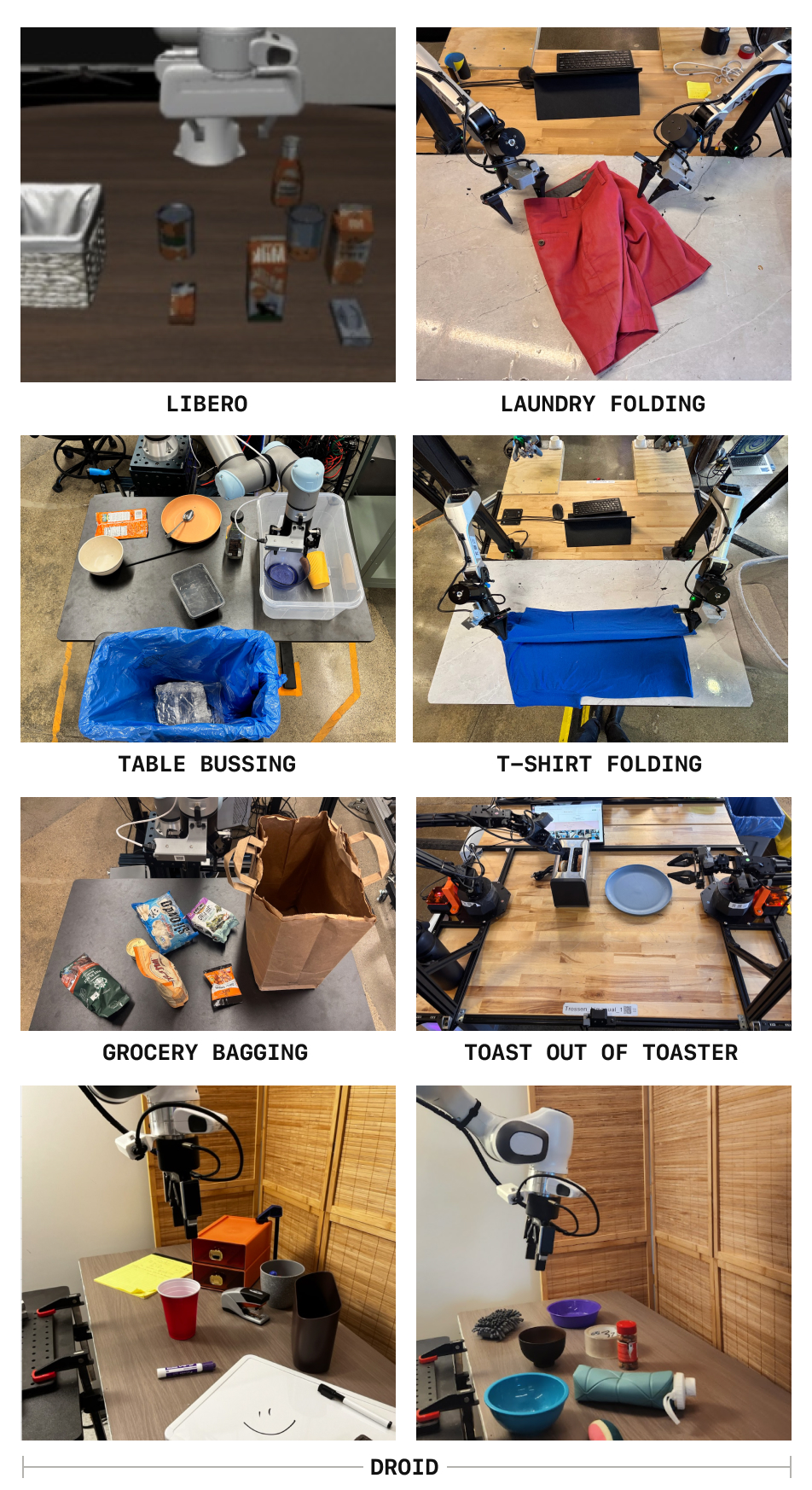}
    \caption{\textbf{Evaluation environments.} We test \ModelAcronym{} across 7~evaluation environments: 6~real-robot tasks and 1~simulation environment. The tasks are designed to test VLA performance on highly dexterous tasks, like folding cloths from a laundry basket (``Laundry Folding''), and generalization, e.g., zero-shot table-top manipulation in unseen environments (``DROID'').
    }
    \label{fig:environments}
\end{figure}

\noindent\textbf{Evaluation tasks.}
We develop a suite of 7~evaluation tasks (6~real robot, 1~simulated; see \cref{fig:environments}), designed to test VLA performance on both, highly dexterous tasks like laundry folding, and generalization tasks, like performing table-top manipulations 0-shot in unseen environments.
\begin{itemize}
    \item \textbf{Libero}: We test on the Libero~\citep{liu2024libero} simulated benchmark suites. We measure average performance across Libero-Spatial, Libero-Object, Libero-Goal, and Libero-10. 
    \item\textbf{Table bussing}~\citep{black2024pi_0} (20~Hz): a UR5 single-arm robot needs to clean a table, sorting 12~objects into a trash bin (for trash) and a plastic container (for plates, bowls, cups and cutlery). The task requires precise grasping of various objects.
    \item\textbf{T-Shirt folding}~\citep{black2024pi_0} (50~Hz): a bi-manual ARX robot setup needs to fold various shirts on a stationary table top. At the beginning of the task, the shirts are placed flat on the table. Succeeding at the task requires precise grasps and movements to fold the shirt.
    \item\textbf{Grocery bagging}~\citep{black2024pi_0} (20~Hz): a UR5 single-arm robot needs to pack seven objects from a table into a grocery bag, taking care to not topple or rip the bag in the process. This task requires picking a diverse set of objects and carefully inserting them into the bag.
    \item\textbf{Toast out of toaster}~\citep{black2024pi_0} (50~Hz): a bimanual Trossen Viper-X robot needs to remove two slices of bread from a toaster and place them on a plate. This task requires precise grasping and placement of the bread slices.
    \item\textbf{Laundry folding}~\citep{black2024pi_0} (50~Hz): a bi-manual ARX robot needs to take shirts and shorts from a basket, flatten them on a table, fold and stack them. This is the most dexterous task we test. It requires precise grasps, dynamic motions to flatten the cloths, retrying and corrections when cloths got tangled up, and precise placements of the folded cloths on the existing stack of cloths. We report success rate on individual clothing items.
    \item\textbf{Zero-shot DROID tabletop manipulation}~\citep{khazatsky2024droid} (15~Hz): we test a policy trained on the full DROID dataset across various table-top manipulation tasks like picking and placing objects, wiping, opening and closing drawers etc. Importantly, we test the policy in a completely \emph{unseen} environment, with a new table setup, background, novel objects, viewpoint and table height. To our knowledge, this is the first ``zero-shot'' evaluation of DROID policies in a completely unseen environment, without co-training or fine-tuning, simply by prompting a pre-trained model with natural language.
\end{itemize}
Following \citet{black2024pi_0}, we use grocery bagging, the toaster task, and laundry folding only to evaluate our most powerful, generalist VLA in \cref{sec:generalist_vlas}. We provide additional details on training datasets and evaluation tasks in \cref{sec:app_exp_task_details}.

\noindent\textbf{Comparisons.}
We test \textbf{\ModelAcronym}, our DCT-based action tokenization approach, trained on each evaluation dataset individually, and \textbf{\ModelUniversalAcronym}, our universal DCT-based action tokenizer, trained on a large dataset of 1M~action sequences. Note that we trained the universal tokenizer on the most diverse real robot dataset we could assemble, which includes data from our real-robot evaluation tasks. We compare both tokenizers to the per-dimension binning scheme used by prior autoregressive VLAs like RT-2~\citep{rt22023arxiv}, RT-2-X~\citep{open_x_embodiment_rt_x_2023} and OpenVLA~\citep{kim2024openvla}, dubbed \textbf{na\"{i}ve tokenization}. We apply the binning tokenization to each time step in the action chunk separately and then concatenate. Finally, while our approach provides a compressed tokenization without the need to train any separate model, we can consider an alternative compression scheme that instead trains a model to produce a quantized representation of the action chunk via \textbf{FSQ}~\cite{fsq}, a simpler alternative to VQ-VAE~\cite{vqvae}. This tokenization strategy has been previously used to tokenize high-dimensional image data~\citep{fsq,yu2023magvit}, and can be viewed as an ablation of our compression-based approach, utilizing compressed representations but with a more complex learning-based alternative to our relatively simple DCT-based method.

\subsection{Comparing Action Tokenizers for VLA Training}
\label{sec:main_comparison}

\begin{table}[h]
\centering
\begin{tabular}{l|c|c|c|c|c}
\toprule
\multirow{2}{*}{Dataset} & \multirow{2}{*}{\begin{tabular}[c]{@{}c@{}}Action\\ Dimension\end{tabular}} & \multirow{2}{*}{\begin{tabular}[c]{@{}c@{}}Control\\ Frequency\end{tabular}} & \multicolumn{2}{c|}{Avg. Token} & \multirow{2}{*}{Compression} \\
 &  &  & Naive & \ModelAcronym &  \\
\midrule
BridgeV2 & 7 & 5\;Hz & 35 & \cellcolor{lightgreen}20 & 1.75 \\
DROID & 7 & 15\;Hz & 105 & \cellcolor{lightgreen}29 & 3.6\\
Bussing & 7 & 20\;Hz & 140 & \cellcolor{lightgreen}28 & 5.0\\
Shirt Fold & 14 & 50\;Hz & 700 & \cellcolor{lightgreen}53 & 13.2\\
\bottomrule
\end{tabular}
\caption{\textbf{Comparison of the average token count per action chunk} for na\"{i}ve tokenization and \ModelAcronym. We use 1-second chunks in all datasets. With our method, each chunk requires many fewer tokens, particularly for high-frequency domains such as the T-shirt folding task, indicating that it is more effective at removing redundancy.}
\label{tab:compression_ratios}
\end{table}

We first provide a comparison of compression rates between our proposed \ModelAcronym~tokenizer and the na\"{i}ve binning scheme used in prior works in \cref{tab:compression_ratios}. We use 1-second action chunks from datasets with various action dimensionalities and control frequencies. For both approaches we use the default hyperparameters, which have comparable tokenization errors. We see that \ModelAcronym~achieves a significant compression of the input action sequences across all datasets. The compression benefits are especially pronounced for datasets with high-frequency action data. Interestingly, \ModelAcronym~consistently generates roughly 30 action tokens per chunk per robot arm (i.e., 60 tokens for the bi-manual setup) in each of the domains. This suggests that \ModelAcronym~finds a representation that approximates the complexity of the underlying action signal, and is largely independent of the frequency of the action data.

We note that this compression is not entirely lossless, with a trade-off between compression ratio and reconstruction accuracy determined by the scale parameter $\gamma$ from \cref{alg:fast}. Figures in \cref{tab:compression_ratios} are at comparable reconstruction accuracy. Please see \cref{sec:app_compression_plots} for plots showing the trade-off between compression and fidelity for each of the tokenizers we compare.

\begin{figure*}[t]
    \centering
    \includegraphics[width=0.8\linewidth]{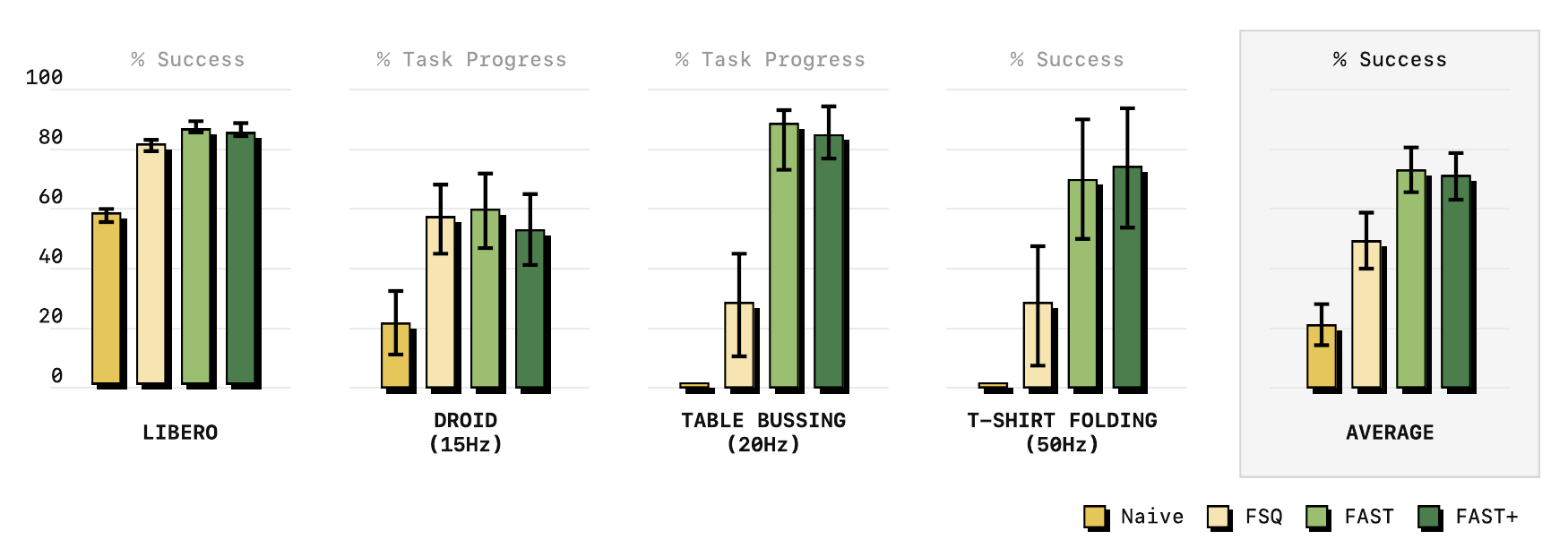}
    \caption{\textbf{Comparison of policy performance using different tokenization approaches.} We find that tokenization approaches that compress action targets (\ModelAcronym, FSQ) lead to substantially more efficient training than the na\"{i}ve binning tokenization used in prior VLAs. Overall, we find that \ModelAcronym~leads to more effective policy training than FSQ, particularly on dexterous real-robot tasks. Our universal tokenizer, \ModelUniversalAcronym, matches the performance of dataset-specific tokenizers. We report mean and 95\% CI.
    }
    \label{fig:tokenization_comparison}
\end{figure*}

Next, we train policies using the policy architecture and tokenization approaches described in \cref{sec:exp_setup}.
We report results in \cref{fig:tokenization_comparison}.

Overall, we find that the na\"{i}ve tokenization applied in prior works struggles to learn effective policies on high-frequency robot data. This is particularly apparent for the highest frequency tasks in our evaluations: Table Bussing (20Hz) and T-Shirt Folding (50Hz). On both tasks, policies trained with na\"{i}ve tokenization are unable to make progress on the task. 

In contrast, we find that compression-based tokenization leads to effective training. Comparing \ModelAcronym{} to our FSQ baseline, we find that \ModelAcronym{} is as good or at times better, particularly on the dexterous, high-frequency tasks, despite being much simpler and requiring no separate neural network training. %

\begin{figure}[t]
    \centering
    \includegraphics[width=\linewidth]{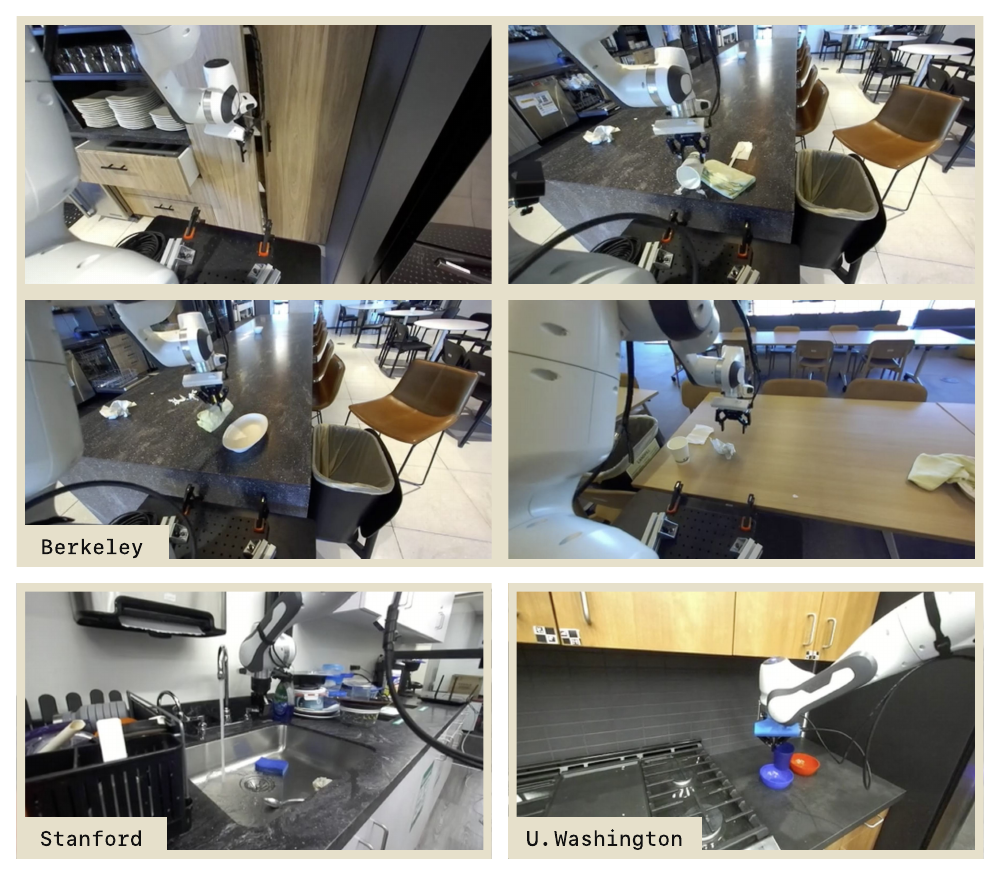}
    \caption{\textbf{Evaluation environments of \ModelAcronym{} policy trained on DROID~\citep{khazatsky2024droid}.} We find that the same policy checkpoint generalizes robustly, and performs various simple table-top tasks \emph{zero-shot} across three university campuses. 
    }
    \label{fig:droid_quali}
\end{figure}

Notably, \ModelAcronym{} tokenization enables the first successful training of a strong generalist policy on the DROID dataset~\citep{khazatsky2024droid}, which can be evaluated \emph{zero-shot} in unseen environments, without fine-tuning, by simply prompting it in natural language. All prior works, including the original DROID paper~\citep{khazatsky2024droid} and OpenVLA~\citep{kim2024openvla}, did not show zero-shot results and focused entirely on co-training or fine-tuning evaluations instead. We demonstrate the generality of our DROID policy by testing it on various table-top manipulation tasks in environments across three university campuses (\cref{fig:droid_quali}).
Out of the box, the policy can competently perform simple manipulation tasks, like picking and placing objects, opening and closing cupboards and turning on faucets, across a wide range of scenes and camera viewpoints. Even unsuccessful trials show sensible behavior, like approaching the handles of microwave and dish washer doors, even if ultimately failing to open them. We show success and failure videos on our website. While far from perfect, the level of generality and robustness of this policy substantially exceeds that of prior DROID policies. %

\subsection{Universal Action Tokenizer}
\label{sec:uniact}

\begin{figure}[t]
    \centering
    \includegraphics[width=\linewidth]{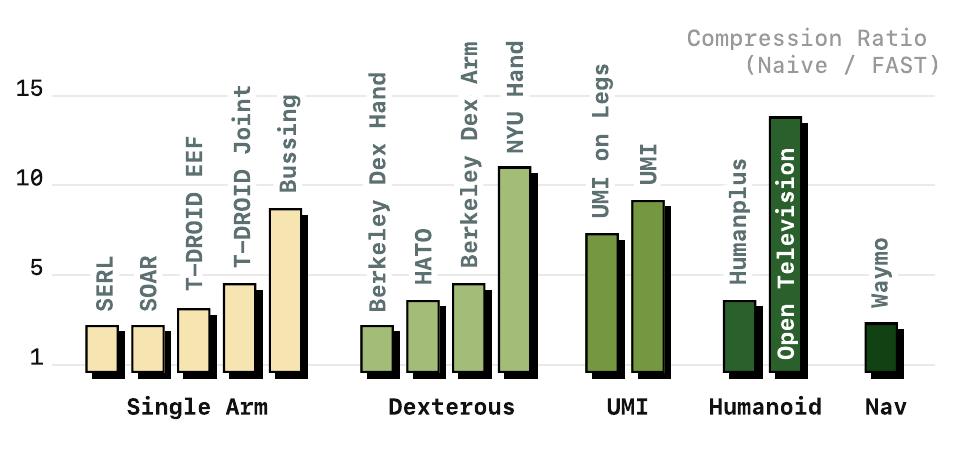}
    \caption{\textbf{Universal tokenizer.} We test the compression rate achieved by our {\ModelUniversalAcronym} tokenizer vs. na\"{i}ve tokenization across diverse robot datasets, \emph{unseen} during tokenizer training. We find that {\ModelAcronym} is effective across a wide range of robot morphologies, action spaces and control frequencies.
    }
    \label{fig:universal_tokenizer_results}
\end{figure}

In this section, we evaluate the performance of our \emph{universal} action tokenizer, \ModelUniversalAcronym, which we trained on 1M real robot action sequences (see \cref{sec:universal_tokenizer}). 
To test the \emph{generality} of the tokenizer, we assemble a diverse set of small testing datasets. This set spans a wide range of robot morphologies, action spaces, and control frequencies (see \cref{fig:universal_tokenizer_results}, with a full list of datasets in \cref{sec:app_univeral_test_set}). Note that none of these datasets is part of the tokenizer training set. They thus test a scenario in which the tokenizer is applied to a completely new robot setup without recomputing the tokenization. We find that the \ModelUniversalAcronym{} tokenizer achieves good compression performance across a wide range of robot datasets, reducing the number of action tokens by 2x across all datasets, and significantly more on some.

We also test performance of the universal tokenizer for policy training, and report results alongside the per-dataset tokenizers in \cref{fig:tokenization_comparison}. 
Across all tasks, the \emph{universal} tokenizer closely matches the performance of the dataset-specific \ModelAcronym~tokenizers, suggesting that the universal tokenizer can be used as a strong default for robot action tokenization.

\subsection{Ablation Studies}

We analyze two key aspects of our method:
(1)~Is our \ModelAcronym~tokenization approach \emph{independent} of the underlying VLA backbone? (2)~How important is the BPE compression step, the only learned component of our tokenization pipeline.

\begin{wrapfigure}{r}{0.4\linewidth}
    \centering
    \vspace{-0.3cm}
    \includegraphics[width=\linewidth]{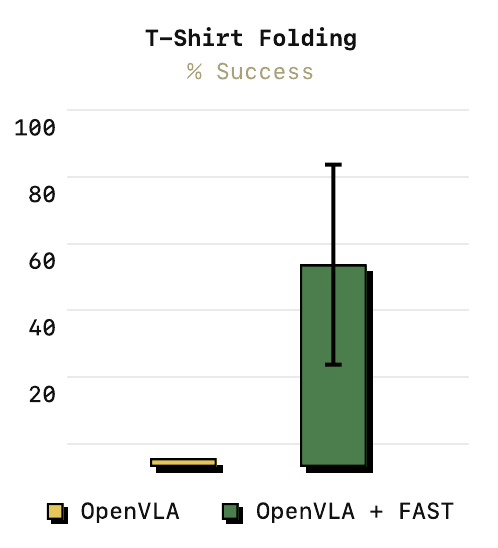}
    \label{fig:openvla_results}
    \vspace{-0.3cm}
\end{wrapfigure}
To answer the first question, we train an OpenVLA policy~\citep{kim2024openvla} on the challenging high-frequency T-shirt folding dataset, comparing the na\"{i}ve tokenization approach originally used in OpenVLA to our \ModelUniversalAcronym{} tokenizer. To comply with the task setup, we modify the OpenVLA model code to accept multiple input images and predict 1-second action chunks. The results on the right demonstrate that \ModelAcronym{} is able to significantly boost performance of OpenVLA, enabling it to train effectively on high-frequency robot manipulation data. This suggests, that our tokenization approach is \emph{independent} of the underlying model backbone, and may be easily applied to a wide range of pre-trained autoregressive transformer models.

\begin{wrapfigure}{r}{0.4\linewidth}
    \centering
    \vspace{-0.3cm}
    \includegraphics[width=\linewidth]{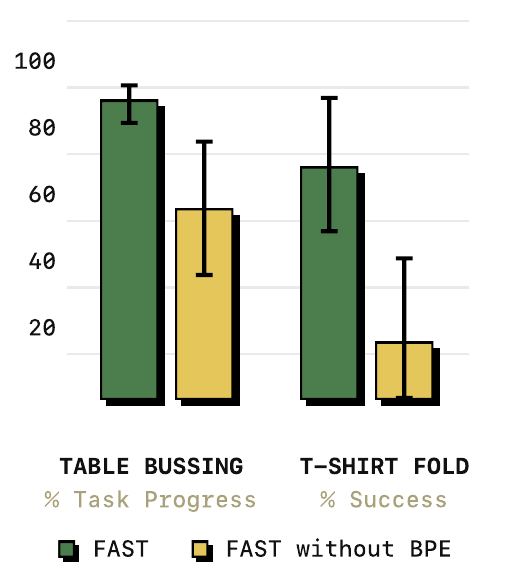}
    \label{fig:openvla_results}
    \vspace{-0.3cm}
\end{wrapfigure}
Secondly, we ablate the BPE encoding step on the table bussing and T-shirt folding tasks. The figure on the right shows that the resulting policies \emph{without BPE encoding} achieve worse rollout performance (but still outperform na\"{i}ve tokenization). Intuitively, the DCT transform still concentrates most of the signal's information in a few tokens, improving the learning signal. However, without BPE, there is a large number of repeated 0-tokens which dilute the learning signal and also significantly slow down inference, since models need to autoregressively predict hundreds of action tokens, ultimately leading to worse policy performance.

\subsection{Comparing \ModelAcronym{} to Diffusion}
\label{sec:ar_vs_diffusion}

In this section, we compare $\pi_0$, a state-of-the-art diffusion VLA, to our model that combines $\pi_0$ with \ModelAcronym{} and uses autoregressive decoding. We compare the performance of both models on the tasks from \cref{sec:main_comparison}. %

\begin{figure}[t]
    \centering
    \includegraphics[width=\linewidth]{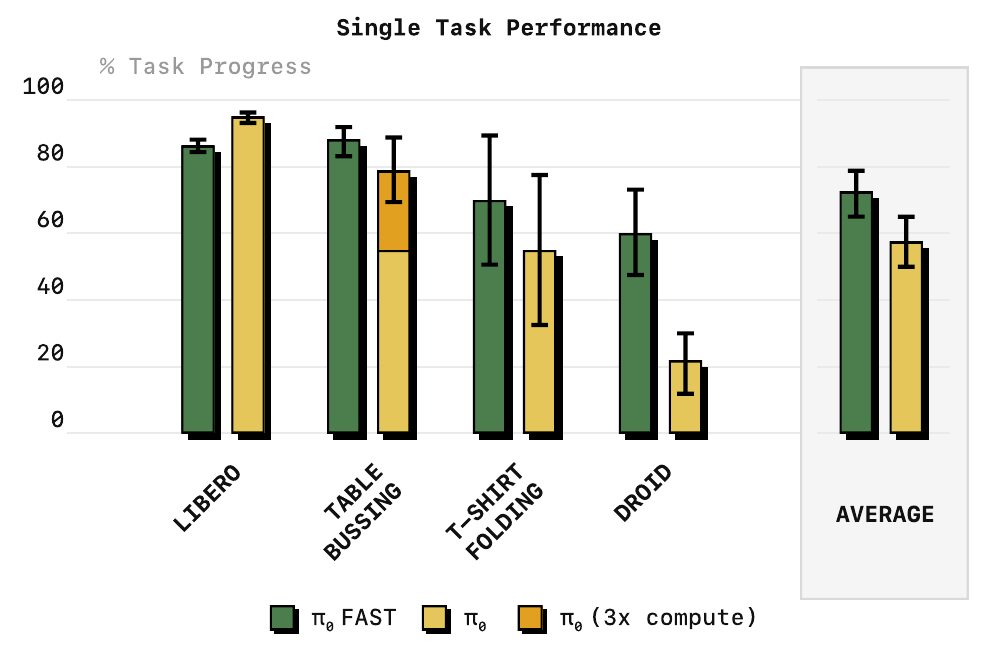}
    \caption{\textbf{Comparison of diffusion $\pi_0$~\citep{black2024pi_0} to our $\pi_0$ model with \ModelAcronym{} decoding on single-task training.} On small datasets (Libero, T-Shirt Folding), both perform comparably. On large datasets (Table Bussing), \ModelAcronym~converges faster. In DROID, we find that \ModelAcronym~follows language instructions better. %
    We report mean and 95\% CI.
    }
    \label{fig:pi0_single_task_comparison}
\end{figure}

We report results in \cref{fig:pi0_single_task_comparison}. We find that on small datasets (Libero, T-Shirt Folding; $<$50h), both VLAs perform comparably. However, on large datasets like Table Bussing, we find that the {\ModelAcronym}-based VLA converges significantly faster, reaching high performance with 3x fewer training steps than the diffusion variant of $\pi_0$.
Additionally, we find that the autoregressive $\pi_0$ model trained with {\ModelAcronym} tokenization follows language instructions more closely: in the DROID evaluations, the diffusion $\pi_0$ model often ignores the language instructions, leading to a lower score. 
We will leave a detailed investigation of the language following abilities of diffusion and autoregressive VLAs to future work.

\begin{figure*}[t]
    \centering
    \includegraphics[width=\linewidth]{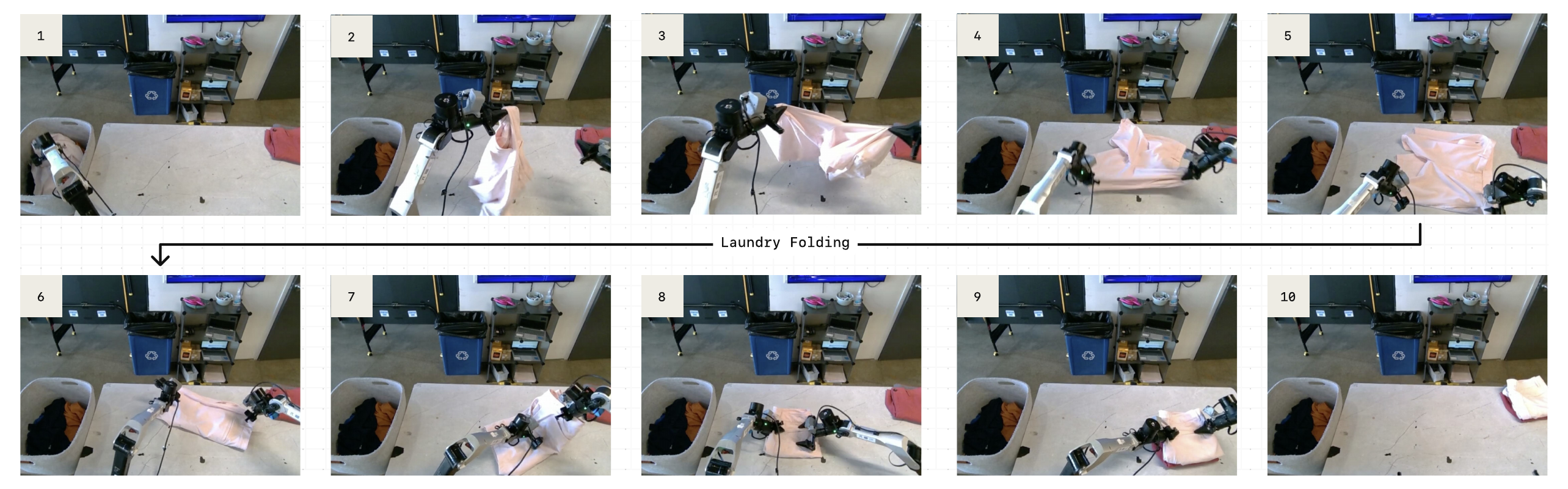}
    \caption{\textbf{Rollout of \GeneralistModelAcronym{} on the laundry folding task.} \ModelAcronym{} tokenization enables autoregressive VLAs to perform complex, long-horizon, and dexterous tasks that were impossible with previous tokenization schemes.
    }
    \label{fig:quali_rollout}
\end{figure*}

One current limitation of the autoregressive VLA is its inference speed: while $\pi_0$ with diffusion typically predicts one second action chunks within 100ms on an NVIDIA 4090 GPU, the $\pi_0$ model with \ModelAcronym{} tokenization needs approximately 750ms of inference time per chunk, since it must perform more autoregressive decoding steps (typically 30-60 action tokens need to be decoded, vs. 10 diffusion steps for diffusion $\pi_0$) and use the full 2B parameter language model backbone for autoregressive decoding (vs. a 300M parameter ``action expert'' for diffusion $\pi_0$). While we did not find this slower inference to hurt performance on the static manipulation tasks we evaluated, it made evaluations significantly slower. Going forward, there are many techniques for accelerating the inference of discrete, autoregressive transformer models that are used extensively in the LLM literature (e.g., speculative decoding, quantization, custom inference kernels, etc.), but we will leave an investigation of these to future work.

\subsection{Scaling Autoregressive VLAs to Large Robot Datasets}
\label{sec:generalist_vlas}

We have demonstrated \ModelAcronym's effectiveness for training autoregressive VLAs on individual robot datasets, but does it scale to training dexterous \emph{generalist} policies? To test this, we train the \GeneralistModelAcronym{} model from the previous section on the cross-embodied robot data mixture used by $\pi_0$~\citep{black2024pi_0}, the largest dexterous robot manipulation dataset to date. It includes 903M timesteps from our own datasets.
Additionally, 9.1\% of the training mixture consists of the open-source datasets BRIDGE v2 \cite{walke2023bridgedata}, DROID \cite{khazatsky2024droid}, and OXE \cite{open_x_embodiment_rt_x_2023}.

\begin{figure}[t]
    \centering
    \includegraphics[width=\linewidth]{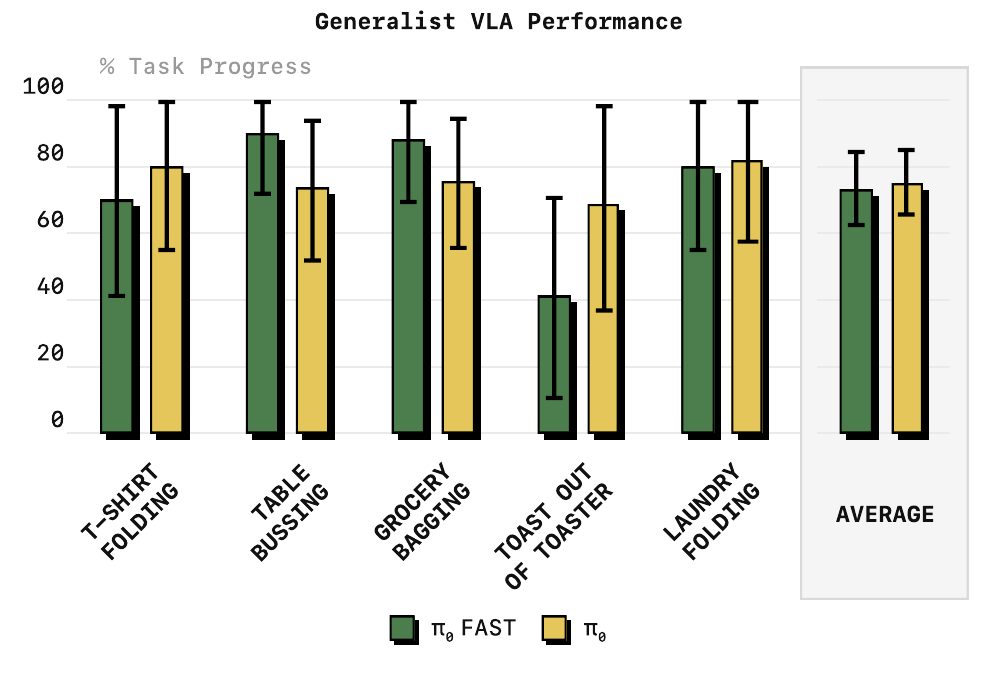}
    \caption{\textbf{Comparison of {\GeneralistModelAcronym} and diffusion $\pi_0$~\citep{black2024pi_0} generalist policies.} \GeneralistModelAcronym{} matches the performance of diffusion $\pi_0$ while requiring significantly less compute for training. Reported: mean and 95\% CI.
    }
    \label{fig:pi0_multi_task_comparison}
\end{figure}
We compare zero-shot performance to the diffusion $\pi_0$ model on the tasks from \citet{black2024pi_0} in \cref{fig:pi0_multi_task_comparison}. Overall, we find that the autoregressive \GeneralistModelAcronym{} model matches the performance of the diffusion $\pi_0$ model, including on the most challenging \textit{laundry folding} task, \textbf{while requiring significantly less compute for training}. We show a qualitative example of \GeneralistModelAcronym{} performing the laundry folding task in \cref{fig:quali_rollout} and include additional videos on our website. 

Importantly, we find that \GeneralistModelAcronym{} converges significantly faster than the diffusion $\pi_0$ model: the model in the evaluations above required 5x fewer GPU hours for training than the $\pi_0$ model from \citet{black2024pi_0}. We show robot evaluation results for multiple checkpoints throughout the course of training in \cref{fig:convergence} (averaging performance on two representative tasks: table bussing and t-shirt folding). The results show clearly that \GeneralistModelAcronym{} achieves high performance significantly faster. For state-of-the-art VLA training runs, which can often use thousands of GPU hours, a 5x reduction in required compute is significant. We include a full comparison across all tasks for a compute-matched $\pi_0$ checkpoint in Appendix, \cref{fig:pi0_compute_matched} and find that the same conclusions hold: \GeneralistModelAcronym{} clearly outperforms compute matched $\pi_0$ due to its faster convergence.

To summarize, we have demonstrated that \ModelAcronym{} tokenization allows us to train autoregressive VLAs on complex, dexterous robot tasks that prior tokenization schemes completely fail on. We have also shown that \ModelAcronym{}, when combined with state-of-the-art VLAs like $\pi_0$, scales to training generalist, cross-embodied policies that rival the performance of the best diffusion VLAs while being significantly faster to train.

\section{Discussion and Future Work}
\label{sec:conclusion} 

In this paper, we introduced \ModelAcronym{}, an efficient action tokenizer for high-frequency robotic control data. \ModelAcronym{} uses the discrete cosine transform (DCT) followed by byte-pair encoding (BPE) to compress action chunks, leading to significantly better compression than existing action tokenizers across a range of robotics domains. Our real-world and simulated VLA experiments show that \ModelAcronym{} leads to dramatically improved performance over the previously used na\"{i}ve action discretization approaches, and outperforms more complex learned tokenization methods based on vector quantization. We also showed that we can train \ModelUniversalAcronym{}, a \emph{universal} action tokenizer, that can serve as a strong default tokenizer for any robot action sequence. Using it, we trained \GeneralistModelAcronym, a dexterous generalist policy that can match performance of state-of-the-art diffusion VLAs, while being significantly more efficient to train. 

There are many exciting directions for future work:

\textbf{Action tokenizers.} While we believe that \ModelAcronym{} is a significant step toward general purpose robot action tokenizers, many questions remain. 
In this work, we tested \ModelAcronym{} on static robot manipulators. Our offline experiments demonstrated promising compression capabilities of \ModelUniversalAcronym{} on other robot morphologies like mobile robots, dexterous hands, and humanoids. Testing actual policy performance on these platforms is an exciting direction for future work. Additionally, exploring alternative compression schemes, and testing the combination of compression-based action encodings with non-autoregressive decoding approaches like diffusion~\citep{black2024pi_0} are interesting directions for future investigation.

\textbf{VLA architectures.} 
Our paper has taken initial steps to explore the trade-offs between two major classes of VLA architectures, autoregressive and diffusion decoding VLAs, but the jury on the best VLA architecture is still out. Future work should carefully explore trade-offs in training speed, language grounding abilities, and expressiveness of either approach. 

\textbf{Inference speed.} While \GeneralistModelAcronym{} matches the overall performance of diffusion $\pi_0$, it is slower at inference time (see \cref{sec:ar_vs_diffusion}). While the slower inference speed was acceptable on the static tasks we evaluated, future work should explore approaches for speeding up inference of autoregressive VLA models to enable them to solve highly dynamic tasks. There is a large literature of inference optimizations for large language models that can be readily applied to autoregressive VLAs.

\section*{Acknowledgements}

We thank Ury Zhilinsky and Kevin Black for their help with setting up data and training infrastructure used in this project. We also thank Pranav Atreya, Haohuan Wang, Lucy Shi, Arhan Jain and Andy Yun for help with DROID policy evaluations at UC Berkeley, Stanford and the University of Washington, and Will Chen for testing and debugging our open-source implementation of {\ModelUniversalAcronym}. We thank Noah Brown, Szymon Jakubczak, Adnan Esmail, Tim Jones, Mohith Mothukuri and James Tanner for help with robot maintenance, and Anna Walling for help with robot, data and eval operations. We are grateful to the whole team of robot operators at Physical Intelligence for their enormous contributions to running data collection and policy evaluations. Finally, we thank Claudio Guglieri, Lachy Groom and Karol Hausman for their help with visualizations used in this paper and on the project website.

\bibliographystyle{plainnat}
\bibliography{references}

\begin{thebibliography}{75}
\providecommand{\natexlab}[1]{#1}
\providecommand{\url}[1]{\texttt{#1}}
\expandafter\ifx\csname urlstyle\endcsname\relax
  \providecommand{\doi}[1]{doi: #1}\else
  \providecommand{\doi}{doi: \begingroup \urlstyle{rm}\Url}\fi

\bibitem[Ahmed et~al.(1974)Ahmed, Natarajan, and Rao]{dct}
Nasir Ahmed, T\_ Natarajan, and Kamisetty~R Rao.
\newblock Discrete cosine transform.
\newblock \emph{IEEE transactions on Computers}, 100\penalty0 (1):\penalty0 90--93, 1974.

\bibitem[Ahn et~al.(2022)Ahn, Brohan, Brown, Chebotar, Cortes, David, Finn, Fu, Gopalakrishnan, Hausman, Herzog, Ho, Hsu, Ibarz, Ichter, Irpan, Jang, Ruano, Jeffrey, Jesmonth, Joshi, Julian, Kalashnikov, Kuang, Lee, Levine, Lu, Luu, Parada, Pastor, Quiambao, Rao, Rettinghouse, Reyes, Sermanet, Sievers, Tan, Toshev, Vanhoucke, Xia, Xiao, Xu, Xu, Yan, and Zeng]{saycan2022arxiv}
Michael Ahn, Anthony Brohan, Noah Brown, Yevgen Chebotar, Omar Cortes, Byron David, Chelsea Finn, Chuyuan Fu, Keerthana Gopalakrishnan, Karol Hausman, Alex Herzog, Daniel Ho, Jasmine Hsu, Julian Ibarz, Brian Ichter, Alex Irpan, Eric Jang, Rosario~Jauregui Ruano, Kyle Jeffrey, Sally Jesmonth, Nikhil Joshi, Ryan Julian, Dmitry Kalashnikov, Yuheng Kuang, Kuang-Huei Lee, Sergey Levine, Yao Lu, Linda Luu, Carolina Parada, Peter Pastor, Jornell Quiambao, Kanishka Rao, Jarek Rettinghouse, Diego Reyes, Pierre Sermanet, Nicolas Sievers, Clayton Tan, Alexander Toshev, Vincent Vanhoucke, Fei Xia, Ted Xiao, Peng Xu, Sichun Xu, Mengyuan Yan, and Andy Zeng.
\newblock Do as i can and not as i say: Grounding language in robotic affordances.
\newblock In \emph{arXiv preprint arXiv:2204.01691}, 2022.

\bibitem[Belkhale and Sadigh(2024)]{belkhale2024minivla}
Suneel Belkhale and Dorsa Sadigh.
\newblock Minivla: A better vla with a smaller footprint, 2024.
\newblock URL \url{https://github.com/Stanford-ILIAD/openvla-mini}.

\bibitem[Belkhale et~al.(2024)Belkhale, Ding, Xiao, Sermanet, Vuong, Tompson, Chebotar, Dwibedi, and Sadigh]{belkhale2024rthactionhierarchiesusing}
Suneel Belkhale, Tianli Ding, Ted Xiao, Pierre Sermanet, Quon Vuong, Jonathan Tompson, Yevgen Chebotar, Debidatta Dwibedi, and Dorsa Sadigh.
\newblock Rt-h: Action hierarchies using language, 2024.
\newblock URL \url{https://arxiv.org/abs/2403.01823}.

\bibitem[Beyer et~al.(2024)Beyer, Steiner, Pinto, Kolesnikov, Wang, Salz, Neumann, Alabdulmohsin, Tschannen, Bugliarello, et~al.]{beyer2024paligemma}
Lucas Beyer, Andreas Steiner, Andr{\'e}~Susano Pinto, Alexander Kolesnikov, Xiao Wang, Daniel Salz, Maxim Neumann, Ibrahim Alabdulmohsin, Michael Tschannen, Emanuele Bugliarello, et~al.
\newblock Paligemma: A versatile 3b vlm for transfer.
\newblock \emph{arXiv preprint arXiv:2407.07726}, 2024.

\bibitem[Bharadhwaj et~al.(2024)Bharadhwaj, Vakil, Sharma, Gupta, Tulsiani, and Kumar]{bharadhwaj2023roboagent}
Homanga Bharadhwaj, Jay Vakil, Mohit Sharma, Abhinav Gupta, Shubham Tulsiani, and Vikash Kumar.
\newblock Roboagent: Generalization and efficiency in robot manipulation via semantic augmentations and action chunking.
\newblock In \emph{2024 IEEE International Conference on Robotics and Automation (ICRA)}, pages 4788--4795. IEEE, 2024.

\bibitem[Black et~al.(2024)Black, Brown, Driess, Esmail, Equi, Finn, Fusai, Groom, Hausman, Ichter, et~al.]{black2024pi_0}
Kevin Black, Noah Brown, Danny Driess, Adnan Esmail, Michael Equi, Chelsea Finn, Niccolo Fusai, Lachy Groom, Karol Hausman, Brian Ichter, et~al.
\newblock $pi\_0 $: A vision-language-action flow model for general robot control.
\newblock \emph{arXiv preprint arXiv:2410.24164}, 2024.

\bibitem[Brohan et~al.(2022{\natexlab{a}})Brohan, Brown, Carbajal, Chebotar, Dabis, Finn, Gopalakrishnan, Hausman, Herzog, Hsu, Ibarz, Ichter, Irpan, Jackson, Jesmonth, Joshi, Julian, Kalashnikov, Kuang, Leal, Lee, Levine, Lu, Malla, Manjunath, Mordatch, Nachum, Parada, Peralta, Perez, Pertsch, Quiambao, Rao, Ryoo, Salazar, Sanketi, Sayed, Singh, Sontakke, Stone, Tan, Tran, Vanhoucke, Vega, Vuong, Xia, Xiao, Xu, Xu, Yu, and Zitkovich]{rt12022arxiv}
Anthony Brohan, Noah Brown, Justice Carbajal, Yevgen Chebotar, Joseph Dabis, Chelsea Finn, Keerthana Gopalakrishnan, Karol Hausman, Alex Herzog, Jasmine Hsu, Julian Ibarz, Brian Ichter, Alex Irpan, Tomas Jackson, Sally Jesmonth, Nikhil Joshi, Ryan Julian, Dmitry Kalashnikov, Yuheng Kuang, Isabel Leal, Kuang-Huei Lee, Sergey Levine, Yao Lu, Utsav Malla, Deeksha Manjunath, Igor Mordatch, Ofir Nachum, Carolina Parada, Jodilyn Peralta, Emily Perez, Karl Pertsch, Jornell Quiambao, Kanishka Rao, Michael Ryoo, Grecia Salazar, Pannag Sanketi, Kevin Sayed, Jaspiar Singh, Sumedh Sontakke, Austin Stone, Clayton Tan, Huong Tran, Vincent Vanhoucke, Steve Vega, Quan Vuong, Fei Xia, Ted Xiao, Peng Xu, Sichun Xu, Tianhe Yu, and Brianna Zitkovich.
\newblock Rt-1: Robotics transformer for real-world control at scale.
\newblock In \emph{arXiv preprint arXiv:2212.06817}, 2022{\natexlab{a}}.

\bibitem[Brohan et~al.(2022{\natexlab{b}})Brohan, Brown, Carbajal, Chebotar, Dabis, Finn, Gopalakrishnan, Hausman, Herzog, Hsu, et~al.]{brohan2022rt}
Anthony Brohan, Noah Brown, Justice Carbajal, Yevgen Chebotar, Joseph Dabis, Chelsea Finn, Keerthana Gopalakrishnan, Karol Hausman, Alex Herzog, Jasmine Hsu, et~al.
\newblock Rt-1: Robotics transformer for real-world control at scale.
\newblock \emph{arXiv preprint arXiv:2212.06817}, 2022{\natexlab{b}}.

\bibitem[Brohan et~al.(2023)Brohan, Brown, Carbajal, Chebotar, Chen, Choromanski, Ding, Driess, Dubey, Finn, Florence, Fu, Arenas, Gopalakrishnan, Han, Hausman, Herzog, Hsu, Ichter, Irpan, Joshi, Julian, Kalashnikov, Kuang, Leal, Lee, Lee, Levine, Lu, Michalewski, Mordatch, Pertsch, Rao, Reymann, Ryoo, Salazar, Sanketi, Sermanet, Singh, Singh, Soricut, Tran, Vanhoucke, Vuong, Wahid, Welker, Wohlhart, Wu, Xia, Xiao, Xu, Xu, Yu, and Zitkovich]{rt22023arxiv}
Anthony Brohan, Noah Brown, Justice Carbajal, Yevgen Chebotar, Xi~Chen, Krzysztof Choromanski, Tianli Ding, Danny Driess, Avinava Dubey, Chelsea Finn, Pete Florence, Chuyuan Fu, Montse~Gonzalez Arenas, Keerthana Gopalakrishnan, Kehang Han, Karol Hausman, Alex Herzog, Jasmine Hsu, Brian Ichter, Alex Irpan, Nikhil Joshi, Ryan Julian, Dmitry Kalashnikov, Yuheng Kuang, Isabel Leal, Lisa Lee, Tsang-Wei~Edward Lee, Sergey Levine, Yao Lu, Henryk Michalewski, Igor Mordatch, Karl Pertsch, Kanishka Rao, Krista Reymann, Michael Ryoo, Grecia Salazar, Pannag Sanketi, Pierre Sermanet, Jaspiar Singh, Anikait Singh, Radu Soricut, Huong Tran, Vincent Vanhoucke, Quan Vuong, Ayzaan Wahid, Stefan Welker, Paul Wohlhart, Jialin Wu, Fei Xia, Ted Xiao, Peng Xu, Sichun Xu, Tianhe Yu, and Brianna Zitkovich.
\newblock Rt-2: Vision-language-action models transfer web knowledge to robotic control.
\newblock In \emph{arXiv preprint arXiv:2307.15818}, 2023.

\bibitem[Cheang et~al.(2024)Cheang, Chen, Jing, Kong, Li, Li, Liu, Wu, Xu, Yang, Zhang, and Zhu]{cheang2024gr2generativevideolanguageactionmodel}
Chi-Lam Cheang, Guangzeng Chen, Ya~Jing, Tao Kong, Hang Li, Yifeng Li, Yuxiao Liu, Hongtao Wu, Jiafeng Xu, Yichu Yang, Hanbo Zhang, and Minzhao Zhu.
\newblock Gr-2: A generative video-language-action model with web-scale knowledge for robot manipulation.
\newblock \emph{arXiv preprint arXiv:2410.06158}, 2024.

\bibitem[Chen et~al.(2022)Chen, Wu, Wang, Liu, Tompkins, Chen, and Wei]{chen2022beats}
Sanyuan Chen, Yu~Wu, Chengyi Wang, Shujie Liu, Daniel Tompkins, Zhuo Chen, and Furu Wei.
\newblock Beats: Audio pre-training with acoustic tokenizers.
\newblock \emph{arXiv preprint arXiv:2212.09058}, 2022.

\bibitem[Cheng et~al.(2024{\natexlab{a}})Cheng, Ji, Yang, Zou, Kautz, Biyik, Yin, Liu, and Wang]{cheng2024navila}
An-Chieh Cheng, Yandong Ji, Zhaojing Yang, Xueyan Zou, Jan Kautz, Erdem Biyik, Hongxu Yin, Sifei Liu, and Xiaolong Wang.
\newblock {NaVILA: Legged Robot Vision-Language-Action Model for Navigation}.
\newblock \emph{arXiv preprint arXiv:2412.04453}, 2024{\natexlab{a}}.

\bibitem[Cheng et~al.(2024{\natexlab{b}})Cheng, Li, Yang, Yang, and Wang]{cheng2024tv}
Xuxin Cheng, Jialong Li, Shiqi Yang, Ge~Yang, and Xiaolong Wang.
\newblock Open-television: Teleoperation with immersive active visual feedback.
\newblock \emph{arXiv preprint arXiv:2407.01512}, 2024{\natexlab{b}}.

\bibitem[Chi et~al.(2023)Chi, Feng, Du, Xu, Cousineau, Burchfiel, and Song]{chi2023diffusionpolicy}
Cheng Chi, Siyuan Feng, Yilun Du, Zhenjia Xu, Eric Cousineau, Benjamin Burchfiel, and Shuran Song.
\newblock Diffusion policy: Visuomotor policy learning via action diffusion.
\newblock In \emph{Proceedings of Robotics: Science and Systems (RSS)}, 2023.

\bibitem[Chi et~al.(2024)Chi, Xu, Pan, Cousineau, Burchfiel, Feng, Tedrake, and Song]{chi2024universal}
Cheng Chi, Zhenjia Xu, Chuer Pan, Eric Cousineau, Benjamin Burchfiel, Siyuan Feng, Russ Tedrake, and Shuran Song.
\newblock Universal manipulation interface: In-the-wild robot teaching without in-the-wild robots.
\newblock In \emph{Proceedings of Robotics: Science and Systems (RSS)}, 2024.

\bibitem[Collaboration et~al.(2023)Collaboration, Padalkar, Pooley, Jain, Bewley, Herzog, Irpan, Khazatsky, Rai, Singh, et~al.]{collaboration2023open}
OX-Embodiment Collaboration, A~Padalkar, A~Pooley, A~Jain, A~Bewley, A~Herzog, A~Irpan, A~Khazatsky, A~Rai, A~Singh, et~al.
\newblock {Open X-Embodiment}: Robotic learning datasets and {RT-X} models.
\newblock \emph{arXiv preprint arXiv:2310.08864}, 1\penalty0 (2), 2023.

\bibitem[Cooley and Tukey(1965)]{fft}
James~W Cooley and John~W Tukey.
\newblock An algorithm for the machine calculation of complex fourier series.
\newblock \emph{Mathematics of computation}, 19\penalty0 (90):\penalty0 297--301, 1965.

\bibitem[Di~Palo and Johns(2024)]{dipalo2024kat}
Norman Di~Palo and Edward Johns.
\newblock Keypoint action tokens enable in-context imitation learning in robotics.
\newblock In \emph{Proceedings of Robotics: Science and Systems (RSS)}, 2024.

\bibitem[Doshi et~al.(2024)Doshi, Walke, Mees, Dasari, and Levine]{Doshi24-crossformer}
Ria Doshi, Homer Walke, Oier Mees, Sudeep Dasari, and Sergey Levine.
\newblock Scaling cross-embodied learning: One policy for manipulation, navigation, locomotion and aviation.
\newblock In \emph{Conference on Robot Learning}, 2024.

\bibitem[Driess et~al.(2023)Driess, Xia, Sajjadi, Lynch, Chowdhery, Ichter, Wahid, Tompson, Vuong, Yu, et~al.]{driess2023palm}
Danny Driess, Fei Xia, Mehdi~SM Sajjadi, Corey Lynch, Aakanksha Chowdhery, Brian Ichter, Ayzaan Wahid, Jonathan Tompson, Quan Vuong, Tianhe Yu, et~al.
\newblock Palm-e: An embodied multimodal language model.
\newblock \emph{arXiv preprint arXiv:2303.03378}, 2023.

\bibitem[Esser et~al.(2020)Esser, Rombach, and Ommer]{esser2020taming}
Patrick Esser, Robin Rombach, and Björn Ommer.
\newblock Taming transformers for high-resolution image synthesis, 2020.

\bibitem[Ettinger et~al.(2021)Ettinger, Cheng, Caine, Liu, Zhao, Pradhan, Chai, Sapp, Qi, Zhou, Yang, Chouard, Sun, Ngiam, Vasudevan, McCauley, Shlens, and Anguelov]{Ettinger_2021_ICCV}
Scott Ettinger, Shuyang Cheng, Benjamin Caine, Chenxi Liu, Hang Zhao, Sabeek Pradhan, Yuning Chai, Ben Sapp, Charles~R. Qi, Yin Zhou, Zoey Yang, Aur'elien Chouard, Pei Sun, Jiquan Ngiam, Vijay Vasudevan, Alexander McCauley, Jonathon Shlens, and Dragomir Anguelov.
\newblock Large scale interactive motion forecasting for autonomous driving: The waymo open motion dataset.
\newblock In \emph{Proceedings of the IEEE/CVF International Conference on Computer Vision (ICCV)}, pages 9710--9719, October 2021.

\bibitem[Fang et~al.(2024{\natexlab{a}})Fang, Fang, Tang, Liu, Wang, Wang, Zhu, and Lu]{fang2024rh20t}
Hao-Shu Fang, Hongjie Fang, Zhenyu Tang, Jirong Liu, Chenxi Wang, Junbo Wang, Haoyi Zhu, and Cewu Lu.
\newblock Rh20t: A comprehensive robotic dataset for learning diverse skills in one-shot.
\newblock In \emph{2024 IEEE International Conference on Robotics and Automation (ICRA)}, pages 653--660. IEEE, 2024{\natexlab{a}}.

\bibitem[Fang et~al.(2024{\natexlab{b}})Fang, Liu, Abbeel, and Levine]{fangandliu2024moka}
Kuan Fang, Fangchen Liu, Pieter Abbeel, and Sergey Levine.
\newblock Moka: Open-world robotic manipulation through mark-based visual prompting.
\newblock \emph{Robotics: Science and Systems (RSS)}, 2024{\natexlab{b}}.

\bibitem[Fu et~al.(2024)Fu, Zhao, Wu, Wetzstein, and Finn]{fu2024humanplus}
Zipeng Fu, Qingqing Zhao, Qi~Wu, Gordon Wetzstein, and Chelsea Finn.
\newblock Humanplus: Humanoid shadowing and imitation from humans.
\newblock In \emph{Conference on Robot Learning ({CoRL})}, 2024.

\bibitem[Gage(1994)]{gage1994new}
Philip Gage.
\newblock A new algorithm for data compression.
\newblock \emph{The C Users Journal}, 12\penalty0 (2):\penalty0 23--38, 1994.

\bibitem[Gillick et~al.(2016)Gillick, Brunk, Vinyals, and Subramanya]{gillick2016bytelm}
Dan Gillick, Cliff Brunk, Oriol Vinyals, and Amarnag Subramanya.
\newblock Multilingual language processing from bytes, 2016.
\newblock URL \url{https://arxiv.org/abs/1512.00103}.

\bibitem[Gong et~al.(2021)Gong, Chung, and Glass]{gong21b_interspeech}
Yuan Gong, Yu-An Chung, and James Glass.
\newblock {AST: Audio Spectrogram Transformer}.
\newblock In \emph{Proc. Interspeech 2021}, pages 571--575, 2021.
\newblock \doi{10.21437/Interspeech.2021-698}.

\bibitem[Guzey et~al.(2024)Guzey, Dai, Savva, Bhirangi, and Pinto]{guzey2024bridging}
Irmak Guzey, Yinlong Dai, Georgy Savva, Raunaq Bhirangi, and Lerrel Pinto.
\newblock Bridging the human to robot dexterity gap through object-oriented rewards, 2024.
\newblock URL \url{https://arxiv.org/abs/2410.23289}.

\bibitem[Ha et~al.(2024)Ha, Gao, Fu, Tan, and Song]{ha2024umilegs}
Huy Ha, Yihuai Gao, Zipeng Fu, Jie Tan, and Shuran Song.
\newblock {UMI} on legs: Making manipulation policies mobile with manipulation-centric whole-body controllers.
\newblock In \emph{Proceedings of the 2024 Conference on Robot Learning}, 2024.

\bibitem[Huang et~al.(2024)Huang, Wang, Li, Zhang, and Fei-Fei]{huang2024rekep}
Wenlong Huang, Chen Wang, Yunzhu Li, Ruohan Zhang, and Li~Fei-Fei.
\newblock Rekep: Spatio-temporal reasoning of relational keypoint constraints for robotic manipulation.
\newblock \emph{arXiv preprint arXiv:2409.01652}, 2024.

\bibitem[Huffman(1952)]{huffmancode}
David~A. Huffman.
\newblock A method for the construction of minimum-redundancy codes.
\newblock \emph{Proceedings of the IRE}, 40\penalty0 (9):\penalty0 1098--1101, 1952.
\newblock \doi{10.1109/JRPROC.1952.273898}.

\bibitem[Jang et~al.(2024)Jang, Yu, Shin, Abbeel, and Seo]{jang2024efficient}
Huiwon Jang, Sihyun Yu, Jinwoo Shin, Pieter Abbeel, and Younggyo Seo.
\newblock Efficient long video tokenization via coordinated-based patch reconstruction.
\newblock \emph{arXiv preprint arXiv:2411.14762}, 2024.

\bibitem[Jiang et~al.(2024)Jiang, Xie, Lin, Xu, Wan, Mandlekar, Fan, and Zhu]{jiang2024dexmimicgen}
Zhenyu Jiang, Yuqi Xie, Kevin Lin, Zhenjia Xu, Weikang Wan, Ajay Mandlekar, Linxi Fan, and Yuke Zhu.
\newblock Dexmimicgen: Automated data generation for bimanual dexterous manipulation via imitation learning.
\newblock \emph{arXiv preprint arXiv:2410.24185}, 2024.

\bibitem[Jones et~al.(2025)Jones, Mees, Sferrazza, Stachowicz, Abbeel, and Levine]{jones2025beyond}
Joshua Jones, Oier Mees, Carmelo Sferrazza, Kyle Stachowicz, Pieter Abbeel, and Sergey Levine.
\newblock Beyond sight: Finetuning generalist robot policies with heterogeneous sensors via language grounding.
\newblock \emph{arXiv preprint arXiv:2501.04693}, 2025.

\bibitem[Karamcheti et~al.(2024)Karamcheti, Nair, Balakrishna, Liang, Kollar, and Sadigh]{karamcheti2024prismatic}
Siddharth Karamcheti, Suraj Nair, Ashwin Balakrishna, Percy Liang, Thomas Kollar, and Dorsa Sadigh.
\newblock Prismatic vlms: Investigating the design space of visually-conditioned language models.
\newblock In \emph{International Conference on Machine Learning (ICML)}, 2024.

\bibitem[Khazatsky et~al.(2024)Khazatsky, Pertsch, Nair, Balakrishna, Dasari, Karamcheti, Nasiriany, Srirama, Chen, Ellis, Fagan, Hejna, Itkina, Lepert, Ma, Miller, Wu, Belkhale, Dass, Ha, Jain, Lee, Lee, Memmel, Park, Radosavovic, Wang, Zhan, Black, Chi, Hatch, Lin, Lu, Mercat, Rehman, Sanketi, Sharma, Simpson, Vuong, Walke, Wulfe, Xiao, Yang, Yavary, Zhao, Agia, Baijal, Castro, Chen, Chen, Chung, Drake, Foster, Gao, Herrera, Heo, Hsu, Hu, Jackson, Le, Li, Lin, Lin, Ma, Maddukuri, Mirchandani, Morton, Nguyen, O'Neill, Scalise, Seale, Son, Tian, Tran, Wang, Wu, Xie, Yang, Yin, Zhang, Bastani, Berseth, Bohg, Goldberg, Gupta, Gupta, Jayaraman, Lim, Malik, Martín-Martín, Ramamoorthy, Sadigh, Song, Wu, Yip, Zhu, Kollar, Levine, and Finn]{khazatsky2024droid}
Alexander Khazatsky, Karl Pertsch, Suraj Nair, Ashwin Balakrishna, Sudeep Dasari, Siddharth Karamcheti, Soroush Nasiriany, Mohan~Kumar Srirama, Lawrence~Yunliang Chen, Kirsty Ellis, Peter~David Fagan, Joey Hejna, Masha Itkina, Marion Lepert, Yecheng~Jason Ma, Patrick~Tree Miller, Jimmy Wu, Suneel Belkhale, Shivin Dass, Huy Ha, Arhan Jain, Abraham Lee, Youngwoon Lee, Marius Memmel, Sungjae Park, Ilija Radosavovic, Kaiyuan Wang, Albert Zhan, Kevin Black, Cheng Chi, Kyle~Beltran Hatch, Shan Lin, Jingpei Lu, Jean Mercat, Abdul Rehman, Pannag~R Sanketi, Archit Sharma, Cody Simpson, Quan Vuong, Homer~Rich Walke, Blake Wulfe, Ted Xiao, Jonathan~Heewon Yang, Arefeh Yavary, Tony~Z. Zhao, Christopher Agia, Rohan Baijal, Mateo~Guaman Castro, Daphne Chen, Qiuyu Chen, Trinity Chung, Jaimyn Drake, Ethan~Paul Foster, Jensen Gao, David~Antonio Herrera, Minho Heo, Kyle Hsu, Jiaheng Hu, Donovon Jackson, Charlotte Le, Yunshuang Li, Kevin Lin, Roy Lin, Zehan Ma, Abhiram Maddukuri, Suvir Mirchandani, Daniel Morton, Tony Nguyen,
  Abigail O'Neill, Rosario Scalise, Derick Seale, Victor Son, Stephen Tian, Emi Tran, Andrew~E. Wang, Yilin Wu, Annie Xie, Jingyun Yang, Patrick Yin, Yunchu Zhang, Osbert Bastani, Glen Berseth, Jeannette Bohg, Ken Goldberg, Abhinav Gupta, Abhishek Gupta, Dinesh Jayaraman, Joseph~J Lim, Jitendra Malik, Roberto Martín-Martín, Subramanian Ramamoorthy, Dorsa Sadigh, Shuran Song, Jiajun Wu, Michael~C. Yip, Yuke Zhu, Thomas Kollar, Sergey Levine, and Chelsea Finn.
\newblock Droid: A large-scale in-the-wild robot manipulation dataset.
\newblock In \emph{Proceedings of Robotics: Science and Systems}, 2024.

\bibitem[Kim et~al.(2024)Kim, Pertsch, Karamcheti, Xiao, Balakrishna, Nair, Rafailov, Foster, Lam, Sanketi, et~al.]{kim2024openvla}
Moo~Jin Kim, Karl Pertsch, Siddharth Karamcheti, Ted Xiao, Ashwin Balakrishna, Suraj Nair, Rafael Rafailov, Ethan Foster, Grace Lam, Pannag Sanketi, et~al.
\newblock Openvla: An open-source vision-language-action model.
\newblock \emph{arXiv preprint arXiv:2406.09246}, 2024.

\bibitem[Lai et~al.()Lai, Huang, and Gershman]{laiaction}
Lucy Lai, Ann~ZX Huang, and Samuel~J Gershman.
\newblock Action chunking as conditional policy compression.

\bibitem[Lee et~al.(2024)Lee, Wang, Etukuru, Kim, Shafiullah, and Pinto]{lee2024behavior}
Seungjae Lee, Yibin Wang, Haritheja Etukuru, H.~Jin Kim, Nur Muhammad~Mahi Shafiullah, and Lerrel Pinto.
\newblock Behavior generation with latent actions.
\newblock \emph{arXiv preprint arXiv:2403.03181}, 2024.

\bibitem[Lin et~al.(2024)Lin, Zhang, Li, Qi, Yi, Levine, and Malik]{lin2024learning}
Toru Lin, Yu~Zhang, Qiyang Li, Haozhi Qi, Brent Yi, Sergey Levine, and Jitendra Malik.
\newblock Learning visuotactile skills with two multifingered hands.
\newblock \emph{arXiv:2404.16823}, 2024.

\bibitem[Liu et~al.(2024)Liu, Zhu, Gao, Feng, Liu, Zhu, and Stone]{liu2024libero}
Bo~Liu, Yifeng Zhu, Chongkai Gao, Yihao Feng, Qiang Liu, Yuke Zhu, and Peter Stone.
\newblock Libero: Benchmarking knowledge transfer for lifelong robot learning.
\newblock \emph{Advances in Neural Information Processing Systems}, 36, 2024.

\bibitem[Liu et~al.(2023)Liu, Li, Wu, and Lee]{liu2023llava}
Haotian Liu, Chunyuan Li, Qingyang Wu, and Yong~Jae Lee.
\newblock Visual instruction tuning.
\newblock In \emph{Advances in Neural Information Processing Systems (NeurIPS)}, 2023.

\bibitem[Loshchilov and Hutter(2017)]{loshchilov2017decoupled}
Ilya Loshchilov and Frank Hutter.
\newblock Decoupled weight decay regularization.
\newblock \emph{arXiv preprint arXiv:1711.05101}, 2017.

\bibitem[Luo et~al.(2024)Luo, Hu, Xu, Tan, Berg, Sharma, Schaal, Finn, Gupta, and Levine]{luo2024serl}
Jianlan Luo, Zheyuan Hu, Charles Xu, You~Liang Tan, Jacob Berg, Archit Sharma, Stefan Schaal, Chelsea Finn, Abhishek Gupta, and Sergey Levine.
\newblock Serl: A software suite for sample-efficient robotic reinforcement learning, 2024.

\bibitem[Mandlekar et~al.(2018)Mandlekar, Zhu, Garg, Booher, Spero, Tung, Gao, Emmons, Gupta, Orbay, et~al.]{mandlekar2018roboturk}
Ajay Mandlekar, Yuke Zhu, Animesh Garg, Jonathan Booher, Max Spero, Albert Tung, Julian Gao, John Emmons, Anchit Gupta, Emre Orbay, et~al.
\newblock Roboturk: A crowdsourcing platform for robotic skill learning through imitation.
\newblock In \emph{Conference on Robot Learning}, pages 879--893. PMLR, 2018.

\bibitem[Mentzer et~al.(2023)Mentzer, Minnen, Agustsson, and Tschannen]{fsq}
Fabian Mentzer, David Minnen, Eirikur Agustsson, and Michael Tschannen.
\newblock Finite scalar quantization: Vq-vae made simple, 2023.
\newblock URL \url{https://arxiv.org/abs/2309.15505}.

\bibitem[Mete et~al.(2024)Mete, Xue, Wilcox, Chen, and Garg]{mete2024questselfsupervisedskillabstractions}
Atharva Mete, Haotian Xue, Albert Wilcox, Yongxin Chen, and Animesh Garg.
\newblock Quest: Self-supervised skill abstractions for learning continuous control, 2024.
\newblock URL \url{https://arxiv.org/abs/2407.15840}.

\bibitem[Nasiriany et~al.(2024)Nasiriany, Xia, Yu, Xiao, Liang, Dasgupta, Xie, Driess, Wahid, Xu, et~al.]{nasirianypivot}
Soroush Nasiriany, Fei Xia, Wenhao Yu, Ted Xiao, Jacky Liang, Ishita Dasgupta, Annie Xie, Danny Driess, Ayzaan Wahid, Zhuo Xu, et~al.
\newblock Pivot: Iterative visual prompting elicits actionable knowledge for vlms.
\newblock In \emph{Forty-first International Conference on Machine Learning}, 2024.

\bibitem[{Octo Model Team} et~al.(2024){Octo Model Team}, Ghosh, Walke, Pertsch, Black, Mees, Dasari, Hejna, Xu, Luo, Kreiman, Tan, Sanketi, Vuong, Xiao, Sadigh, Finn, and Levine]{octo_2023}
{Octo Model Team}, Dibya Ghosh, Homer Walke, Karl Pertsch, Kevin Black, Oier Mees, Sudeep Dasari, Joey Hejna, Charles Xu, Jianlan Luo, Tobias Kreiman, {You Liang} Tan, Pannag Sanketi, Quan Vuong, Ted Xiao, Dorsa Sadigh, Chelsea Finn, and Sergey Levine.
\newblock Octo: An open-source generalist robot policy.
\newblock In \emph{Proceedings of Robotics: Science and Systems}, Delft, Netherlands, 2024.

\bibitem[{Open X-Embodiment Collaboration} et~al.(2023){Open X-Embodiment Collaboration}, Padalkar, Pooley, Jain, Bewley, Herzog, Irpan, Khazatsky, Rai, Singh, Brohan, Raffin, Wahid, Burgess-Limerick, Kim, Schölkopf, Ichter, Lu, Xu, Finn, Xu, Chi, Huang, Chan, Pan, Fu, Devin, Driess, Pathak, Shah, Büchler, Kalashnikov, Sadigh, Johns, Ceola, Xia, Stulp, Zhou, Sukhatme, Salhotra, Yan, Schiavi, Su, Fang, Shi, Amor, Christensen, Furuta, Walke, Fang, Mordatch, Radosavovic, Leal, Liang, Kim, Schneider, Hsu, Bohg, Bingham, Wu, Wu, Luo, Gu, Tan, Oh, Malik, Tompson, Yang, Lim, Silvério, Han, Rao, Pertsch, Hausman, Go, Gopalakrishnan, Goldberg, Byrne, Oslund, Kawaharazuka, Zhang, Majd, Rana, Srinivasan, Chen, Pinto, Tan, Ott, Lee, Tomizuka, Du, Ahn, Zhang, Ding, Srirama, Sharma, Kim, Kanazawa, Hansen, Heess, Joshi, Suenderhauf, Palo, Shafiullah, Mees, Kroemer, Sanketi, Wohlhart, Xu, Sermanet, Sundaresan, Vuong, Rafailov, Tian, Doshi, Martín-Martín, Mendonca, Shah, Hoque, Julian, Bustamante, Kirmani, Levine, Moore,
  Bahl, Dass, Song, Xu, Haldar, Adebola, Guist, Nasiriany, Schaal, Welker, Tian, Dasari, Belkhale, Osa, Harada, Matsushima, Xiao, Yu, Ding, Davchev, Zhao, Armstrong, Darrell, Jain, Vanhoucke, Zhan, Zhou, Burgard, Chen, Wang, Zhu, Li, Lu, Chebotar, Zhou, Zhu, Xu, Wang, Bisk, Cho, Lee, Cui, hua Wu, Tang, Zhu, Li, Iwasawa, Matsuo, Xu, and Cui]{open_x_embodiment_rt_x_2023}
{Open X-Embodiment Collaboration}, Abhishek Padalkar, Acorn Pooley, Ajinkya Jain, Alex Bewley, Alex Herzog, Alex Irpan, Alexander Khazatsky, Anant Rai, Anikait Singh, Anthony Brohan, Antonin Raffin, Ayzaan Wahid, Ben Burgess-Limerick, Beomjoon Kim, Bernhard Schölkopf, Brian Ichter, Cewu Lu, Charles Xu, Chelsea Finn, Chenfeng Xu, Cheng Chi, Chenguang Huang, Christine Chan, Chuer Pan, Chuyuan Fu, Coline Devin, Danny Driess, Deepak Pathak, Dhruv Shah, Dieter Büchler, Dmitry Kalashnikov, Dorsa Sadigh, Edward Johns, Federico Ceola, Fei Xia, Freek Stulp, Gaoyue Zhou, Gaurav~S. Sukhatme, Gautam Salhotra, Ge~Yan, Giulio Schiavi, Hao Su, Hao-Shu Fang, Haochen Shi, Heni~Ben Amor, Henrik~I Christensen, Hiroki Furuta, Homer Walke, Hongjie Fang, Igor Mordatch, Ilija Radosavovic, Isabel Leal, Jacky Liang, Jaehyung Kim, Jan Schneider, Jasmine Hsu, Jeannette Bohg, Jeffrey Bingham, Jiajun Wu, Jialin Wu, Jianlan Luo, Jiayuan Gu, Jie Tan, Jihoon Oh, Jitendra Malik, Jonathan Tompson, Jonathan Yang, Joseph~J. Lim, João
  Silvério, Junhyek Han, Kanishka Rao, Karl Pertsch, Karol Hausman, Keegan Go, Keerthana Gopalakrishnan, Ken Goldberg, Kendra Byrne, Kenneth Oslund, Kento Kawaharazuka, Kevin Zhang, Keyvan Majd, Krishan Rana, Krishnan Srinivasan, Lawrence~Yunliang Chen, Lerrel Pinto, Liam Tan, Lionel Ott, Lisa Lee, Masayoshi Tomizuka, Maximilian Du, Michael Ahn, Mingtong Zhang, Mingyu Ding, Mohan~Kumar Srirama, Mohit Sharma, Moo~Jin Kim, Naoaki Kanazawa, Nicklas Hansen, Nicolas Heess, Nikhil~J Joshi, Niko Suenderhauf, Norman~Di Palo, Nur Muhammad~Mahi Shafiullah, Oier Mees, Oliver Kroemer, Pannag~R Sanketi, Paul Wohlhart, Peng Xu, Pierre Sermanet, Priya Sundaresan, Quan Vuong, Rafael Rafailov, Ran Tian, Ria Doshi, Roberto Martín-Martín, Russell Mendonca, Rutav Shah, Ryan Hoque, Ryan Julian, Samuel Bustamante, Sean Kirmani, Sergey Levine, Sherry Moore, Shikhar Bahl, Shivin Dass, Shuran Song, Sichun Xu, Siddhant Haldar, Simeon Adebola, Simon Guist, Soroush Nasiriany, Stefan Schaal, Stefan Welker, Stephen Tian, Sudeep Dasari,
  Suneel Belkhale, Takayuki Osa, Tatsuya Harada, Tatsuya Matsushima, Ted Xiao, Tianhe Yu, Tianli Ding, Todor Davchev, Tony~Z. Zhao, Travis Armstrong, Trevor Darrell, Vidhi Jain, Vincent Vanhoucke, Wei Zhan, Wenxuan Zhou, Wolfram Burgard, Xi~Chen, Xiaolong Wang, Xinghao Zhu, Xuanlin Li, Yao Lu, Yevgen Chebotar, Yifan Zhou, Yifeng Zhu, Ying Xu, Yixuan Wang, Yonatan Bisk, Yoonyoung Cho, Youngwoon Lee, Yuchen Cui, Yueh hua Wu, Yujin Tang, Yuke Zhu, Yunzhu Li, Yusuke Iwasawa, Yutaka Matsuo, Zhuo Xu, and Zichen~Jeff Cui.
\newblock Open {X-E}mbodiment: Robotic learning datasets and {RT-X} models.
\newblock \url{https://arxiv.org/abs/2310.08864}, 2023.

\bibitem[Pagnoni et~al.(2024)Pagnoni, Pasunuru, Rodriguez, Nguyen, Muller, Li, Zhou, Yu, Weston, Zettlemoyer, Ghosh, Lewis, Holtzman†, and Iyer]{meta_blt}
Artidoro Pagnoni, Ram Pasunuru, Pedro Rodriguez, John Nguyen, Benjamin Muller, Margaret Li, Chunting Zhou, Lili Yu, Jason Weston, Luke Zettlemoyer, Gargi Ghosh, Mike Lewis, Ari Holtzman†, and Srinivasan Iyer.
\newblock Byte latent transformer: Patches scale better than tokens.
\newblock 2024.
\newblock URL \url{https://github.com/facebookresearch/blt}.

\bibitem[Qi et~al.(2022)Qi, Kumar, Calandra, Ma, and Malik]{qi2022inhand}
Haozhi Qi, Ashish Kumar, Roberto Calandra, Yi~Ma, and Jitendra Malik.
\newblock In-hand object rotation via rapid motor adaptation, 2022.
\newblock URL \url{https://arxiv.org/abs/2210.04887}.

\bibitem[Radford et~al.(2019)Radford, Wu, Child, Luan, Amodei, and Sutskever]{radford2019language}
Alec Radford, Jeff Wu, Rewon Child, David Luan, Dario Amodei, and Ilya Sutskever.
\newblock Language models are unsupervised multitask learners.
\newblock 2019.

\bibitem[Reed et~al.(2022)Reed, Zolna, Parisotto, Colmenarejo, Novikov, Barth-maron, Gim{\'e}nez, Sulsky, Kay, Springenberg, et~al.]{reed2022generalist}
Scott Reed, Konrad Zolna, Emilio Parisotto, Sergio~G{\'o}mez Colmenarejo, Alexander Novikov, Gabriel Barth-maron, Mai Gim{\'e}nez, Yury Sulsky, Jackie Kay, Jost~Tobias Springenberg, et~al.
\newblock A generalist agent.
\newblock \emph{Transactions on Machine Learning Research}, 2022.

\bibitem[Sennrich et~al.(2015)Sennrich, Haddow, and Birch]{sennrich2015neural}
Rico Sennrich, Barry Haddow, and Alexandra Birch.
\newblock Neural machine translation of rare words with subword units.
\newblock \emph{arXiv preprint arXiv:1508.07909}, 2015.

\bibitem[Singh et~al.(2024)Singh, Loquercio, Sferrazza, Wu, Qi, Abbeel, and Malik]{singh2024hop}
Himanshu~Gaurav Singh, Antonio Loquercio, Carmelo Sferrazza, Jane Wu, Haozhi Qi, Pieter Abbeel, and Jitendra Malik.
\newblock Hand-object interaction pretraining from videos, 2024.
\newblock URL \url{https://arxiv.org/abs/2409.08273}.

\bibitem[van~den Oord et~al.(2018)van~den Oord, Vinyals, and Kavukcuoglu]{vqvae}
Aaron van~den Oord, Oriol Vinyals, and Koray Kavukcuoglu.
\newblock Neural discrete representation learning, 2018.
\newblock URL \url{https://arxiv.org/abs/1711.00937}.

\bibitem[Walke et~al.(2023)Walke, Black, Zhao, Vuong, Zheng, Hansen-Estruch, He, Myers, Kim, Du, et~al.]{walke2023bridgedata}
Homer~Rich Walke, Kevin Black, Tony~Z Zhao, Quan Vuong, Chongyi Zheng, Philippe Hansen-Estruch, Andre~Wang He, Vivek Myers, Moo~Jin Kim, Max Du, et~al.
\newblock {BridgeData} v2: A dataset for robot learning at scale.
\newblock In \emph{Conference on Robot Learning}, pages 1723--1736. PMLR, 2023.

\bibitem[Wallace(1992)]{jpeg}
Gregory~K Wallace.
\newblock The jpeg still picture compression standard.
\newblock \emph{IEEE transactions on consumer electronics}, 38\penalty0 (1):\penalty0 xviii--xxxiv, 1992.

\bibitem[Wang et~al.(2024)Wang, Chen, Zhao, and He]{wangscaling}
Lirui Wang, Xinlei Chen, Jialiang Zhao, and Kaiming He.
\newblock Scaling proprioceptive-visual learning with heterogeneous pre-trained transformers.
\newblock In \emph{The Thirty-eighth Annual Conference on Neural Information Processing Systems}, 2024.

\bibitem[Wen et~al.(2024)Wen, Zhu, Li, Zhu, Wu, Xu, Liu, Cheng, Shen, Peng, Feng, and Tang]{wen2024tinyvlafastdataefficientvisionlanguageaction}
Junjie Wen, Yichen Zhu, Jinming Li, Minjie Zhu, Kun Wu, Zhiyuan Xu, Ning Liu, Ran Cheng, Chaomin Shen, Yaxin Peng, Feifei Feng, and Jian Tang.
\newblock Tinyvla: Towards fast, data-efficient vision-language-action models for robotic manipulation.
\newblock \emph{arXiv preprint arXiv:2409.12514}, 2024.

\bibitem[Yan et~al.(2024)Yan, Zaharia, Mnih, Abbeel, Faust, and Liu]{yan2024elastictok}
Wilson Yan, Matei Zaharia, Volodymyr Mnih, Pieter Abbeel, Aleksandra Faust, and Hao Liu.
\newblock Elastictok: Adaptive tokenization for image and video.
\newblock \emph{arXiv preprint arXiv:2410.08368}, 2024.

\bibitem[Ye et~al.(2024)Ye, Jang, Jeon, Joo, Yang, Peng, Mandlekar, Tan, Chao, Lin, et~al.]{ye2024latent}
Seonghyeon Ye, Joel Jang, Byeongguk Jeon, Sejune Joo, Jianwei Yang, Baolin Peng, Ajay Mandlekar, Reuben Tan, Yu-Wei Chao, Bill~Yuchen Lin, et~al.
\newblock Latent action pretraining from videos.
\newblock \emph{arXiv preprint arXiv:2410.11758}, 2024.

\bibitem[Yu et~al.(2023)Yu, Cheng, Sohn, Lezama, Zhang, Chang, Hauptmann, Yang, Hao, Essa, and Jiang]{yu2023magvit}
Lijun Yu, Yong Cheng, Kihyuk Sohn, José Lezama, Han Zhang, Huiwen Chang, Alexander~G. Hauptmann, Ming-Hsuan Yang, Yuan Hao, Irfan Essa, and Lu~Jiang.
\newblock Magvit: Masked generative video transformer, 2023.
\newblock URL \url{https://arxiv.org/abs/2212.05199}.

\bibitem[Zawalski et~al.(2024)Zawalski, Chen, Pertsch, Mees, Finn, and Levine]{Zawalski24-ecot}
Michał Zawalski, William Chen, Karl Pertsch, Oier Mees, Chelsea Finn, and Sergey Levine.
\newblock Robotic control via embodied chain-of-thought reasoning.
\newblock In \emph{Conference on Robot Learning}, 2024.

\bibitem[Zeghidour et~al.(2021)Zeghidour, Luebs, Omran, Skoglund, and Tagliasacchi]{zeghidour2021soundstreamendtoendneuralaudio}
Neil Zeghidour, Alejandro Luebs, Ahmed Omran, Jan Skoglund, and Marco Tagliasacchi.
\newblock Soundstream: An end-to-end neural audio codec, 2021.
\newblock URL \url{https://arxiv.org/abs/2107.03312}.

\bibitem[Zhao et~al.(2023)Zhao, Kumar, Levine, and Finn]{zhao2023learning}
Tony~Z Zhao, Vikash Kumar, Sergey Levine, and Chelsea Finn.
\newblock Learning fine-grained bimanual manipulation with low-cost hardware.
\newblock \emph{arXiv preprint arXiv:2304.13705}, 2023.

\bibitem[Zhao et~al.(2024)Zhao, Tompson, Driess, Florence, Ghasemipour, Finn, and Wahid]{zhao2024alohaunleashedsimplerecipe}
Tony~Z Zhao, Jonathan Tompson, Danny Driess, Pete Florence, Kamyar Ghasemipour, Chelsea Finn, and Ayzaan Wahid.
\newblock Aloha unleashed: A simple recipe for robot dexterity.
\newblock \emph{arXiv preprint arXiv:2410.13126}, 2024.

\bibitem[Zhen et~al.(2024{\natexlab{a}})Zhen, Qiu, Chen, Yang, Yan, Du, Hong, and Gan]{zhen20243d}
Haoyu Zhen, Xiaowen Qiu, Peihao Chen, Jincheng Yang, Xin Yan, Yilun Du, Yining Hong, and Chuang Gan.
\newblock 3d-vla: A 3d vision-language-action generative world model.
\newblock \emph{arXiv preprint arXiv:2403.09631}, 2024{\natexlab{a}}.

\bibitem[Zhen et~al.(2024{\natexlab{b}})Zhen, Qiu, Chen, Yang, Yan, Du, Hong, and Gan]{zhen20243dvla}
Haoyu Zhen, Xiaowen Qiu, Peihao Chen, Jincheng Yang, Xin Yan, Yilun Du, Yining Hong, and Chuang Gan.
\newblock 3d-vla: 3d vision-language-action generative world model.
\newblock \emph{arXiv preprint arXiv:2403.09631}, 2024{\natexlab{b}}.

\bibitem[Zheng et~al.(2024)Zheng, Liang, Huang, Gao, Daum{\'e}~III, Kolobov, Huang, and Yang]{zheng2024tracevla}
Ruijie Zheng, Yongyuan Liang, Shuaiyi Huang, Jianfeng Gao, Hal Daum{\'e}~III, Andrey Kolobov, Furong Huang, and Jianwei Yang.
\newblock Tracevla: Visual trace prompting enhances spatial-temporal awareness for generalist robotic policies.
\newblock \emph{arXiv preprint arXiv:2412.10345}, 2024.

\bibitem[Zhou et~al.(2024)Zhou, Atreya, Lee, Walke, Mees, and Levine]{zhou2024soar}
Zhiyuan Zhou, Pranav Atreya, Abraham Lee, Homer Walke, Oier Mees, and Sergey Levine.
\newblock Autonomous improvement of instruction following skills via foundation models.
\newblock In \emph{Conference on Robot Learning}, 2024.

\bibitem[Ziv and Lempel(1978)]{lempelziv}
Jacob Ziv and Abraham Lempel.
\newblock Compression of individual sequences via variable-rate coding.
\newblock \emph{IEEE transactions on Information Theory}, 24\penalty0 (5):\penalty0 530--536, 1978.

\end{thebibliography}

\clearpage
\appendix

\subsection{Data Mixture for Training Universal Tokenizer}
\label{sec:app_universal_data_mix}

The training mixture for the universal tokenizer mainly consists of the $\pi_0$~\citep{black2024pi_0} datasets described in Section \ref{sec:generalist_vlas}.
For many datasets, we include versions with multiple action space parametrizations: joint space, end-effector world frame, and end-effector camera frame, to ensure the generality of the resulting tokenizer.
Open X-Embodiment~\citep{open_x_embodiment_rt_x_2023}, DROID~\citep{khazatsky2024droid}, and Bridge\;V2~\citep{walke2023bridgedata} are included in their original form. Before tokenization, all actions are padded to 32 dimensions to accommodate action spaces of different dimensionality.

\begin{table}[h]
    \centering
    \resizebox{\linewidth}{!}{
    \begin{tabular}{lcccc}
        \toprule
        Dataset Name & Morphology & Action Space & \makecell{Control \\Frequency\\ (Hz)}& \makecell{Mixture \\Weight\\ (\%)}\\
        \midrule
        ARX & Bi-manual & Joint & 50 & 7.2 \\
        AgileX & Bi-manual & Joint & 50 & 1.8 \\
        Fibocom & Mobile & Joint & 50 & 2.9 \\
        Franka FR3 & Single arm & Joint & 20 & 3.7 \\
        Mobile Trossen & Mobile & Joint & 50 & 2.5 \\
        Trossen Biarm & Bi-manual & Joint & 50 & 4.3 \\
        UR5 single & Single arm & Joint & 20 & 10.3 \\
        UR5 biarm & Bi-manual & Joint & 20 & 2.4 \\
        ARX slate mobile & Mobile & Joint & 50 & 2.5 \\
        \midrule
        ARX EE & Bi-manual & EE & 50 & 3.6 \\
        AgileX EE & Bi-manual & EE & 50 & 0.9 \\
        Fibocom EE & Mobile & EE & 50 & 1.4 \\
        Franka FR3 EE & Single arm & EE & 20 & 1.9 \\
        Mobile Trossen EE & Mobile & EE & 50 & 1.2 \\
        Trossen Biarm EE & Bi-manual & EE & 50 & 2.1 \\
        UR5 single EE & Single arm & EE & 20 & 5.2 \\
        UR5 biarm EE & Bi-manual & EE & 20 & 1.2 \\
        ARX slate mobile EE & Mobile & EE & 50 & 1.2 \\
        \midrule
        ARX Cam & Bi-manual & CamFrame & 50 & 3.6 \\
        AgileX Cam & Bi-manual & CamFrame & 50 & 0.9 \\
        Fibocom Cam & Mobile & CamFrame & 50 & 1.4 \\
        Franka FR3 Cam & Single arm & CamFrame & 20 & 1.9 \\
        Mobile Trossen Cam & Mobile & CamFrame & 50 & 1.2 \\
        Trossen Biarm Cam & Bi-manual & CamFrame & 50 & 2.1 \\
        UR5 single Cam & Single arm & CamFrame & 20 & 5.2 \\
        UR5 biarm Cam & Bi-manual & CamFrame & 20 & 1.2 \\
        ARX slate mobile Cam & Mobile & CamFrame & 50 & 1.2 \\
        \midrule
        ALOHA~\citep{zhao2023learning} & Bi-manual & Joint & 50 & 5.0 \\
        DROID~\citep{khazatsky2024droid} & Single arm & Joint & 15 & 11.2 \\
        Bridge\;V2~\citep{walke2023bridgedata} & Single arm & EE & 5 & 5.0 \\
        OpenX~\citep{open_x_embodiment_rt_x_2023} & Single arm & EE & mixed & 3.8 \\ 
        \bottomrule
    \end{tabular}
    }
\end{table}

\subsection{Trading off Between Compression and Reconstruction}
\label{sec:app_compression_plots}
\begin{figure}[t]
    \centering
    \includegraphics[width=\linewidth]{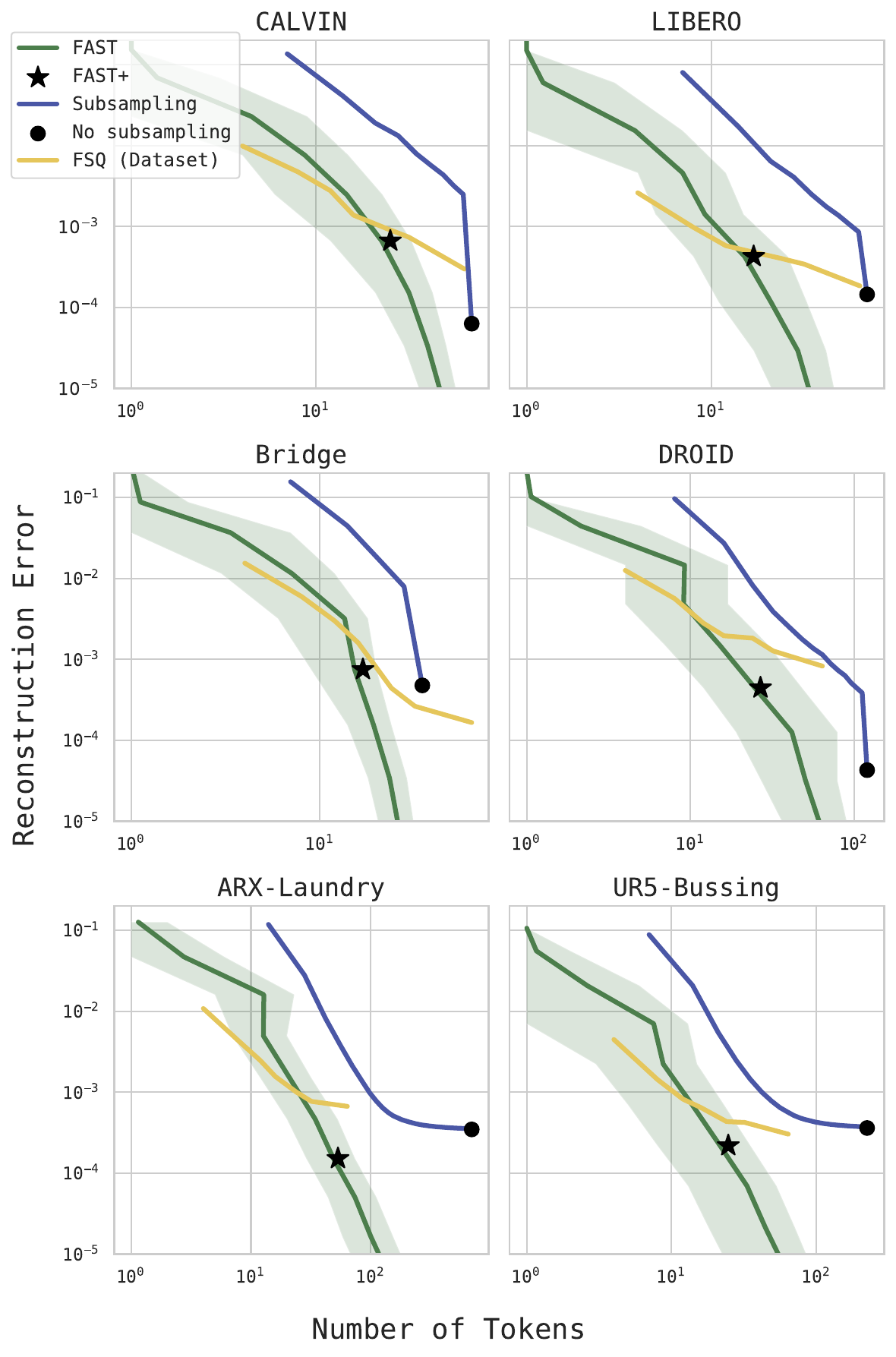}
    \caption{Comparison of compression-reconstruction tradeoff on six training datsets. Any discretization method includes some hyperparameter that controls the tradeoff between reconstruction fidelity and compression level, represented here as number of tokens in the output (vocab size is held constant across all tokenizers). We sweep this hyperparameter (\ModelAcronym: rounding scale; na\"{i}ve tokenization: subsampling frequency; FSQ: number of latent tokens) and find that {\ModelAcronym} performs well across a wide range of scales. In particular, although it is less efficient than VQ-based tokenizers at low fidelities, it exhibits much better scaling to higher reconstruction fidelity, making {\ModelAcronym} much more applicable to fine-grained control problems. Specific instantiations of each tokenizer ({\ModelUniversalAcronym}, and na\"{i}ve tokenization without subsampling) are also shown.}
    \label{fig:dataset-comparison}
\end{figure}

\subsection{Policy Training}
\label{sec:app_policy_training_details}

We train policies with $\pi_0$~\citep{black2024pi_0} and OpenVLA~\citep{kim2024openvla} backbones. Depending on the task, policies are conditioned on two or three inputs images (one third person camera, and one wrist camera per robot arm), using a resolution of 224x224 pixels. The VLA backbones encode each image separately via the pre-trained vision encoder and concatenate the resulting tokens. We additionally condition on a natural language task instruction and the robot's proprioceptive state. Both get tokenized via the LLMs language tokenizer, treating them as strings. For the proprioceptive state, we apply a bin tokenization pre-processing, akin to RT-2's action tokenization~\citep{rt22023arxiv}, discretizing into 256 bins. We then tokenize the integers as part of the text input sequence. Note that a simple bin tokenization scheme is sufficient for the proprioceptive state, since it is an \emph{input} to the policy (as opposed to the action \emph{outputs}, that require advanced tokenization as our experiments demonstrate).

We train all policies using a short linear learning rate warm-up (1k steps) and then a constant learning rate of 5e-5. We use the AdamW optimizer~\citep{loshchilov2017decoupled} ($b1 = 0.9$, $b2 = 0.95$) without weight decay, clip gradient magnitude to 1 and compute an EMA of the network weights with weight 0.999.

During inference, we use simple greedy autoregressive decoding, except for the bi-manual robot tasks (T-shirt folding, toast out of toaster, laundry folding), where we found a small temperature of $\beta = 0.7$ to be helpful to get policies to move out of the home position (since some of the data included stationary chunks of actions where the robot hovers in the initial position at the beginning of training episodes).

\subsection{DROID Policy Setup}
\label{sec:app_droid_exp_details}

Here, we provide further details about our DROID training setup to make it easy for others to reproduce and build on our results. For training on the DROID dataset, we condition the policy on a single third-person view and the wrist camera view. Since DROID provides two external camera views per episode, we randomly sample the third-person view during training. Similarly, DROID provides three natural language annotations for each training episode, and we randomize over them during training. We do not use the camera calibration information. Thus, the trained policy can be tested on new viewpoints out of the box, without the need for calibration. We use joint velocity and absolute gripper position action space, and train the policy to predict 15-step action chunks (we execute 8 or 15-step chunks open-loop at inference time). We apply light data curation: we train only on the episodes marked as ``success'' (75k episodes) and filter out any idle timesteps with all-zero actions during training (usually timesteps in which the teleoperators reset the position of the VR controller during data collection). Other than that, we found training on the full dataset to work well, though there is likely potential for improving performance with more careful curation. We train policies for three epochs (240k iterations @ 256 batch size), which takes approximately 4~days on 8xH100 GPUs for the 3B parameter VLAs we are using.

\subsection{Evaluation Tasks and Training Datasets}
\label{sec:app_exp_task_details}

\begin{figure}[t]
    \centering
    \begin{subfigure}{\linewidth}
        \centering
        \includegraphics[width=\linewidth]{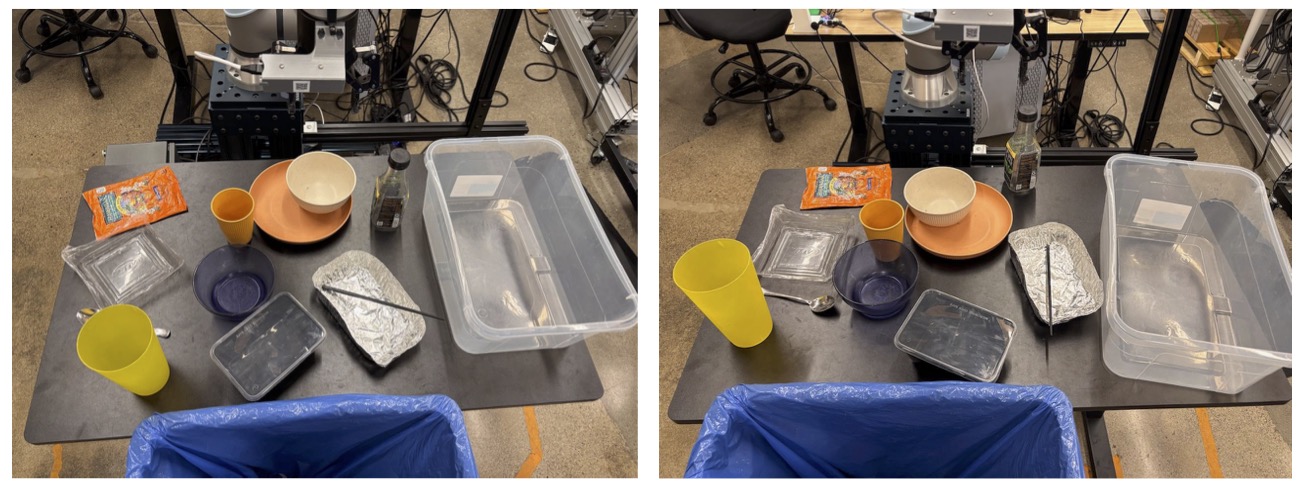}
        \caption{Table Bussing}
        \label{fig:experimental_setup_bus}
    \end{subfigure}
    \begin{subfigure}{\linewidth}
        \centering
        \includegraphics[width=\linewidth]{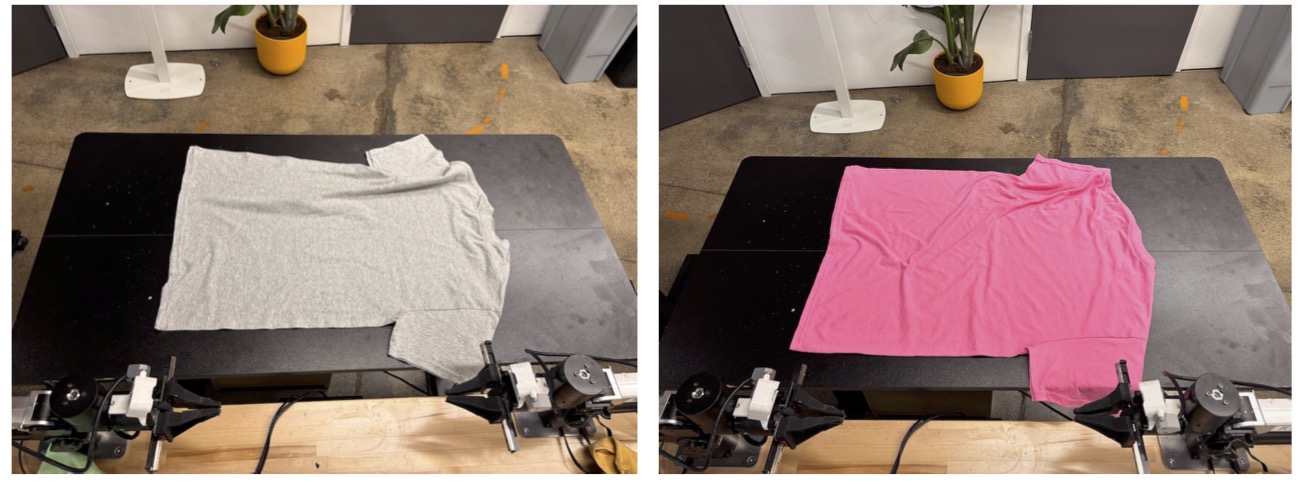}
        \caption{T-Shirt Folding}
        \label{fig:experimental_setup_shirt}
    \end{subfigure}
    \begin{subfigure}{\linewidth}
        \centering
        \includegraphics[width=\linewidth]{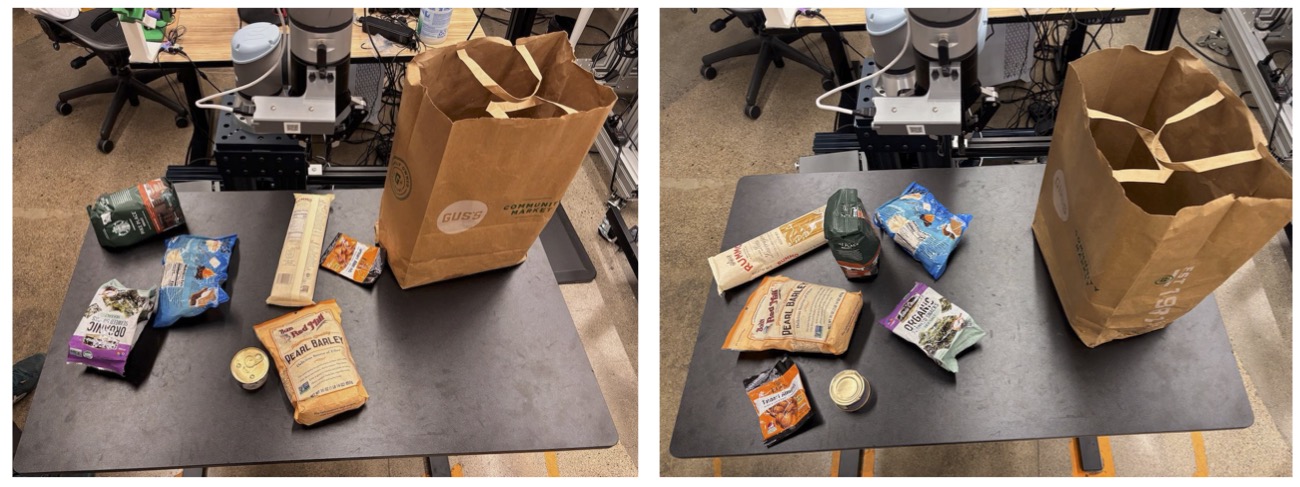}
        \caption{Grocery Bagging}
        \label{fig:experimental_setup_grocery}
    \end{subfigure}
    \begin{subfigure}{\linewidth}
        \centering
        \includegraphics[width=\linewidth]{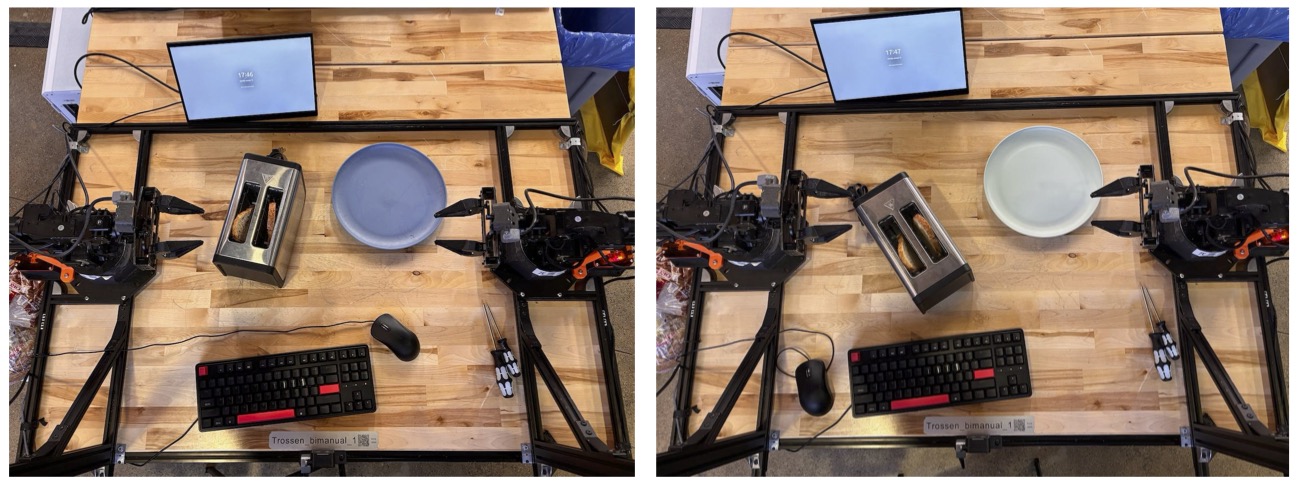}
        \caption{Toast out of Toaster}
        \label{fig:experimental_setup_toast}
    \end{subfigure}
    \begin{subfigure}{\linewidth}
        \centering
        \includegraphics[width=\linewidth]{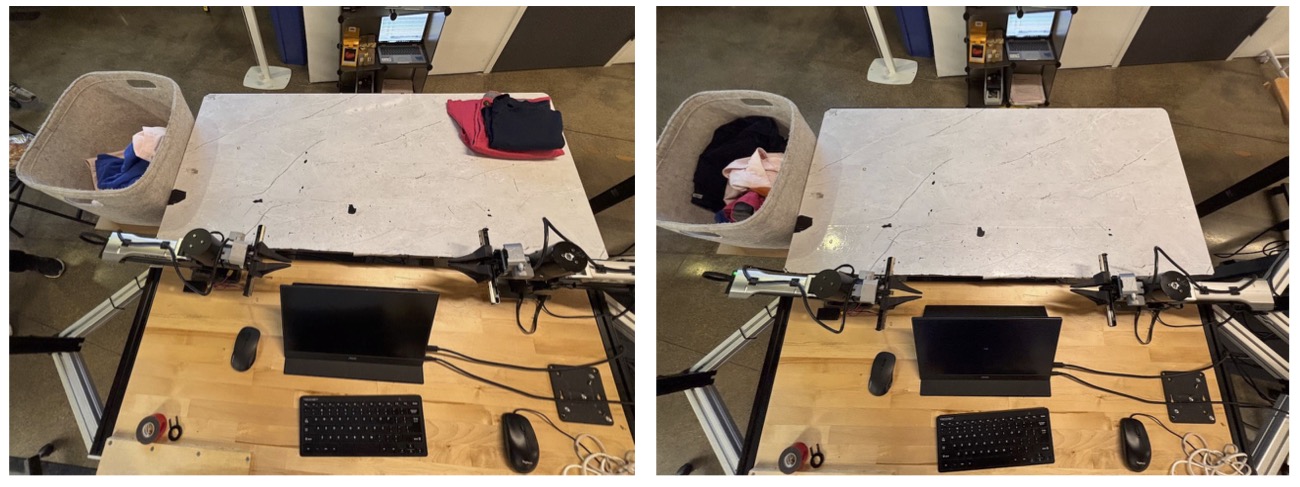}
        \caption{Laundry Folding}
        \label{fig:experimental_setup_laundry}
    \end{subfigure}
    \caption{Sampled initial configurations of evaluation tasks.}
    \label{fig:experimental_setup}
\end{figure}

Below, we describe all evaluation tasks and training datasets used in our experiments. We detail the distribution of initial conditions and scoring criteria.

\textbf{Libero.} We follow the training and evaluation setup of \citet{liu2024libero}. We evaluate on the Libero-Spatial, Libero-Object, Libero-Goal and Libero-Long benchmarking suites and use the corresponding datasets provided by the authors for training. We combine all datasets into one dataset with 270k samples, and train one policy jointly on all to reduce the number of policies that need to be trained. We train all policies for a total of 40k iterations ($\approx40$ epochs). We use the re-rendered datasets of \citet{kim2024openvla} for our experiments. Success is evaluated as a binary criterion per episode.

\textbf{Table Bussing.} 
This task requires a single UR5e robot arm to clean a table by bussing objects (a mixture of trash, plates, and dishes) into a trash can or bussing bin. 
The training dataset contains demonstrations in randomized bussing scenes with approximately 70 objects. 
The evaluation scene, shown in Figure~\ref{fig:experimental_setup_bus}, contains twelve objects on a table in an unseen configuration.
The scene was created to stress the capability of the model, with utensils intentionally placed on top of trash, objects obstructing each other, and challenging objects such as chopsticks, transparent plastic, and reflective containers.
The overall score is calculated as the percentage of objects correctly thrown away or placed in the bin. 

\textbf{T-Shirt Folding.} 
This task requires a bimanual ARX robot to fold a t-shirt.
The training dataset has demonstrations of shirt folding with approximately 150 shirts, varying in size, color, and style.
The evaluation scene, shown in Figure~\ref{fig:experimental_setup_shirt}, cycles through five seen shirts of varying colors and sizes, each starting from a flat configuration.
The overall score is calculated as the percentage of shirts successfully folded, as determined by a human rater.

\textbf{Grocery Bagging.}
This task requires a single UR5e robot arm to bag groceries. 
This task was evaluated out-of-the-box on models pretrained with the full mixture detailed in \citet{black2024pi_0}.
The evaluation scene, shown in Figure~\ref{fig:experimental_setup_grocery}, contains seven items (with varying shapes, sizes, materials, and weights) and a large paper grocery bag. 
The overall score is calculated as the percentage of items placed into the grocery bag. 

\textbf{Toast out of Toaster.}
This task requires a bi-manual Trossen ViperX robot, mirroring the ALOHA~\cite{zhao2024alohaunleashedsimplerecipe} setup, to take two pieces of toast out of a toaster and place them onto a plate. 
This task was evaluated out-of-the-box on models pretrained with the full mixture detailed in \citet{black2024pi_0}.
The evaluation scene is shown in Figure~\ref{fig:experimental_setup_toast} and the overall score tracks task progress, with one point for removing each piece of toast and one point for placing it on the plate, for a score out of four. 

\textbf{Laundry Folding.}
This task requires a bi-manual ARX robot to take a piece of clothing, short or t-shirt, out of a laundry bin and fold it. It is a very challenging task, since successful folding of the tangled up laundry requires multiple steps of unfurling and flattening the laundry before folding can start.
Following \citet{black2024pi_0}, his task was evaluated with models pretrained on the full $\pi_0$ training mixture detailed in \citet{black2024pi_0} and fine-tuned with a small amount of high-quality, task-specific data.
The evaluation scene, shown in Figure~\ref{fig:experimental_setup_laundry}, contains five items of clothing randomly placed in a laundry hamper. 
The overall score is calculated as the percentage of clothing successfully folded and stacked, as determined by a human rater.

\begin{figure}[t]
    \centering
    \includegraphics[width=\linewidth]{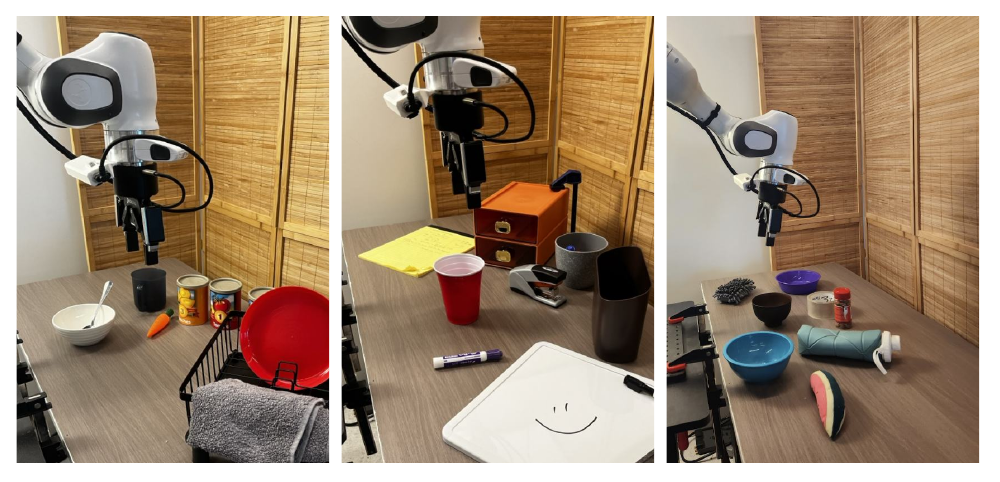}
    \caption{Setups used for quantitative DROID evaluation.
    }
    \label{fig:app_droid_eval_setups}
\end{figure}

\textbf{DROID.} We train on all successful episodes from the DROID dataset (75k episodes, 21M samples) for 240k iterations ($\approx$3 episodes). We apply light data curation (see \cref{sec:app_droid_exp_details}). After training, we deploy the policy \emph{zero-shot} in new scenes, with unseen scene background, camera angles, and objects. For quantitative evaluation, we design an evaluation suite with 16 tasks and 44 trials total per policy (see \cref{tab:droid_eval_tasks}). Each trial is scored with a task progress rubric (e.g., 1 point for picking up the correct object, 1 point for placing it in the correct receptacle). We show example scenes from the quantitative evaluation in \cref{fig:app_droid_eval_setups}. We further run qualitative tests of the policy across various real-world setups on three different university campuses (see \cref{fig:droid_quali}). We do not measure success rates during these evaluations, but provide numerous qualitative videos of successes and failures to help readers get a sense of the policy's capabilities.

\begin{table}[t]
    \centering
    \caption{DROID evaluation tasks.}
    \label{tab:droid_eval_tasks}
    \begin{tabular}{l|c}
        \toprule
        Task & Trials\\
        \midrule
        Put the spoon in the dish rack & 4\\
        Put carrot in bowl & 4\\
        Put plate in dish rack & 2\\
        Wipe the table & 2\\
        Put the plate on the table & 2\\
        Clean up the table & 2\\
        Close the drawer & 4\\
        Put the stapler on the notebook & 2\\
        Put stapler in the drawer & 4\\
        Clean the whiteboard & 2\\
        Put the marker in the cup & 4\\
        Put the black sponge in the blue bowl & 2\\
        Put the red bottle in the black bowl & 2\\
        Put the watermelon in the purple bowl & 2\\
        Move the watermelon from the purple bowl to the blue bowl & 2\\
        Put the tape in the purple bowl & 2\\
        Put the water bottle on the left side of the table & 2\\
        \midrule
        \textbf{Total} & \textbf{44}\\
        \bottomrule
    \end{tabular}
\end{table}

\begin{table*}[]
\renewcommand{\arraystretch}{1.5}
\caption{Universal Tokenizer Evaluation Datasets.}
\label{sec:app_univeral_test_set}
\resizebox{\textwidth}{!}{%
\begin{tabular}{ccccccc}
\hline
  Morphology &
  Dataset Name &
  Platform &
  Action Space &
  Action Dim &
  Control Frequency &
  Task \\ \hline
  \multicolumn{1}{c|}{\multirow{5}{*}{Single Arm}} &
  SOAR~\cite{zhou2024soar} &
  WidowX &
  EEF &
  7 &
  5 &
  Pick/place \\
  \multicolumn{1}{c|}{} &
  DROID-Eval EEF~\cite{khazatsky2024droid} &
  Franka &
  EEF &
  7 &
  15 &
  Pick/place \\
  \multicolumn{1}{c|}{} &
  DROID-Eval Joint~\cite{khazatsky2024droid} &
  Franka &
  Joint &
  8 &
  15 &
  Pick/place \\
  \multicolumn{1}{c|}{} &
  SERL~\cite{luo2024serl} &
  Franka &
  EEF &
  7 &
  10 &
  Insertion \\
  \multicolumn{1}{c|}{} &
  $\pi$ Table Bussing~\cite{black2024pi_0} &
  UR5 &
  Joint &
  8 &
  20 &
  Pick/place \\ \hline %
  \multicolumn{1}{c|}{\multirow{4}{*}{Dexterous}} &
  NYU DexHand~\cite{guzey2024bridging} &
  ALLEGRO &
  Joint+EEF &
  30 &
  16 &
  Dexterous manipulation \\
  \multicolumn{1}{c|}{} &
  Berkeley DexHand~\cite{qi2022inhand} &
  ALLEGRO &
  Joint &
  16 &
  20 &
  In-hand manipulation \\ %
  \multicolumn{1}{c|}{} &
  Berkeley DexArm~\cite{singh2024hop} &
  xArm+ALLEGRO &
  Joint &
  23 &
  20 &
  Dextrous pick/place \\ %
  \multicolumn{1}{c|}{} &
  HATO~\cite{lin2024learning} &
  UR5+Psyonic Hand &
  EEF+Joint &
  24 &
  10 &
  Dextrous pick/place \\ \hline %
  \multicolumn{1}{c|}{\multirow{2}{*}{UMI}} &
  UMI~\cite{chi2024universal} &
  UMI &
  EEF &
  7 &
  20 &
  Pick/place \\
  \multicolumn{1}{c|}{} &
  UMI on Legs~\cite{ha2024umilegs} &
  UMI &
  EEF &
  7 &
  20 &
  Whole-body manipulation \\ \hline %
  \multicolumn{1}{c|}{\multirow{2}{*}{Humanoid}} &
  HumanPlus~\cite{fu2024humanplus} &
  Unitree H1 &
  Joint &
  40 &
  50 &
  Whole-body manipulation \\
  \multicolumn{1}{c|}{} &
  UCSD TeleVision~\cite{cheng2024tv} &
  Unitree H1 w/Neck &
  Joint &
  28 &
  60 &
  Manipulation+active perception \\ \hline
  \multicolumn{1}{c|}{\multirow{1}{*}{Navigation}} &
  Waymo~\cite{Ettinger_2021_ICCV} &
  Waymo Car &
  2D delta &
  2 &
  10 &
  Autonomous Driving\\ \hline
\end{tabular}%
}
\end{table*}

\begin{figure}[t]
    \centering
    \includegraphics[width=\linewidth]{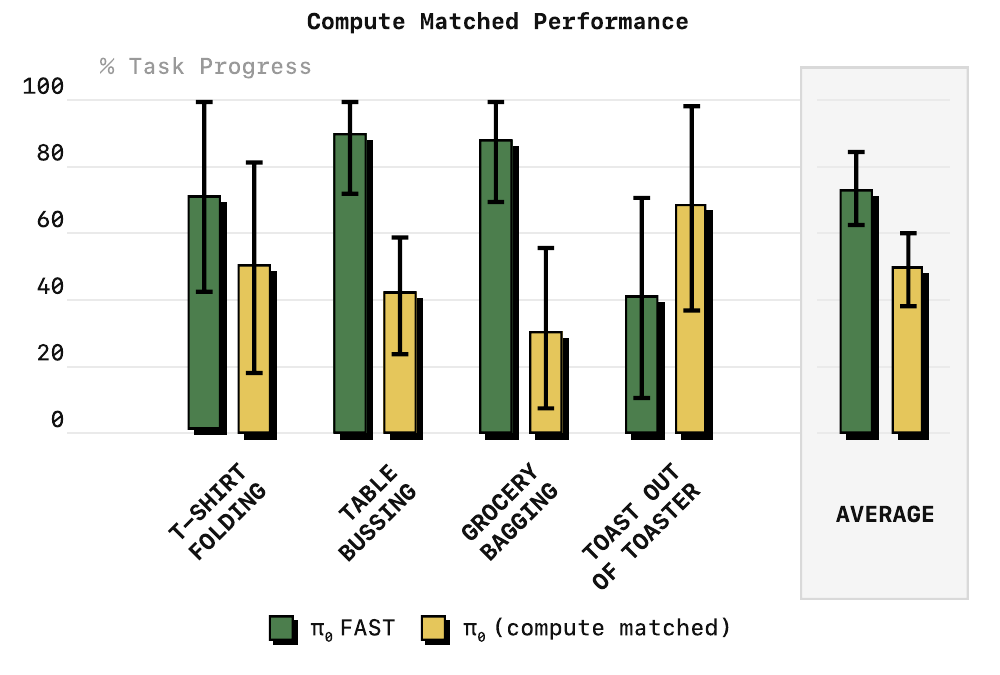}
    \caption{\textbf{Comparison of {\GeneralistModelAcronym} and \emph{compute-matched} diffusion $\pi_0$~\citep{black2024pi_0} generalist policies.} \GeneralistModelAcronym{} clearly outperforms the diffusion VLA when trained with the same amount of training compute, due to its faster convergence. Reported: mean and 95\% CI.
    }
    \label{fig:pi0_compute_matched}
\end{figure}

\end{document}